\documentclass[twocolumn]{article} 

\usepackage{arxiv}

\usepackage{times}
\usepackage{epsfig}
\usepackage{graphicx}
\usepackage{amsmath}
\usepackage{amssymb}

\usepackage{array}

\usepackage{multirow}

\usepackage{url}

\usepackage[accsupp]{axessibility}  

\newtheorem{mydefi}{Definition}
\newcommand{\R}{\mathbb{R}}

\usepackage{hyperref}

\begin{document}
\suppressfloats
\title{Mesh Convolutional Autoencoder for Semi-Regular Meshes of Different Sizes}

\author{Sara Hahner \\
        Fraunhofer Center for Machine Learning and SCAI\\
        {\tt\small sara.hahner@scai.fraunhofer.de}\\
		\And  Jochen Garcke \\
	    Fraunhofer Center for Machine Learning and SCAI\\
	    University of Bonn\\
		{\tt\small jochen.garcke@scai.fraunhofer.de}\\
}

\newcommand{\thedate}{October 20, 2021} 

\twocolumn[
\maketitle


\begin{abstract}
The analysis of deforming 3D surface meshes is accelerated by autoencoders since the low-dimensional embeddings can be used to visualize underlying dynamics.
But, state-of-the-art mesh convolutional autoencoders require a fixed connectivity of all input meshes handled by the autoencoder. This is due to either the use of spectral convolutional layers or mesh dependent pooling operations. 
Therefore, the types of datasets that one can study are limited and the learned knowledge cannot be transferred to other datasets that exhibit similar behavior.
To address this, we transform the discretization of the surfaces to semi-regular meshes that have a locally regular connectivity and whose meshing is hierarchical. This allows us to apply the same spatial convolutional filters to the local neighborhoods and to define a pooling operator that can be applied to every semi-regular mesh. 
We apply the same mesh autoencoder to different datasets and our reconstruction error is more than 50\% lower than the error from state-of-the-art models, which have to be trained for every mesh separately.
Additionally, we visualize the underlying dynamics of unseen mesh sequences with an autoencoder trained on different classes of meshes. 

\end{abstract}
\vspace{4ex}
]

\section{Introduction}

We study three-dimensional data that is discretized by a triangular surface mesh. In particular, we study the deformation of surfaces, which discretize human bodies, animals, or work pieces from computer aided engineering.
Surface deformation is locally described by the same physical rules, which motivates the application of convolution to learn translation-invariant localized features.

\begin{figure}
    \begin{minipage}[c]{\linewidth}
    \centering
    \includegraphics[width=0.35\textwidth, trim={0.3cm 0cm 0.09cm 0.28cm},clip]{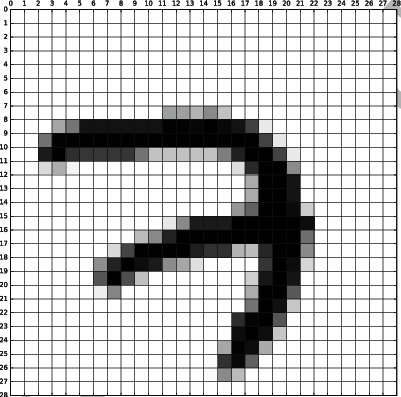}
    \end{minipage}
    \begin{minipage}[c]{\textwidth}
    \centering
    \end{minipage}
    \caption{MNIST dataset sample \cite{LeCun1998} with a visualization of the underlying grid defined by the pixels aligned along the two axes.} 
    \label{gridstructure}
\end{figure}

Convolutional neural networks (CNN) are successful in the analysis and generation of data, especially images, because of their efficient calculation of translation-invariant localized features by sliding filters over the images \cite{LeCun1989}. 
Regular pixel grids describe the 2D images. 
This global grid structure, determined by the two axes of the two-dimensional space (see Figure \ref{gridstructure}), is essential for CNNs, because it implies properties such as a common system of coordinates, shift invariance, and a fixed neighborhood structure \cite{Bronstein2017}. These characteristics allow for an efficient application of the local kernels, the sliding of the kernels along the two axes, and a constant definition of the pooling operator.

While two-dimensional surfaces embedded in $\R^3$ are locally homeomorphic to the two-dimensional space, the surfaces are of non-Euclidean nature.
Therefore, they generally lack the global grid structure, 
which is so essential for the efficient application of CNNs. 
Furthermore, the meshes are usually heterogeneous in the number of vertices, faces, their connectivity, and size, which hinders the direct application of 2D-convolution.

In our novel approach, we calculate an alternative discrete approximation of the surface data based on semi-regular meshes.
Semi-regular meshes have 
regular regional patches, which means that every vertex inside the patch has exactly six neighbors. Also semi-regular meshes have by definition a multi-scale structure that allows us to look at the meshes in different resolutions \cite{Payan2015}.

To this semi-regular approximation of our data, we apply spatial convolution, which follows the idea of 2D-convolution and defines kernels on local neighborhoods of the vertices \cite{Bronstein2017}. 
Since the neighborhoods of most of the vertices of the semi-regular mesh are regular, we use convolutional kernels that can be straightforwardly implemented and preserve the orientation of the neighborhood as well as the permutation of neighbors.
Additionally, the multi-scale structure of the semi-regular mesh allows us to define a general pooling operation that works on all semi-regular meshes.
Since the CNNs learn local features, we feed the regional patches separately to the network. This allows us to apply the network to meshes of different sizes and topology. 
The global context is not lost, but fed to the network via padding. 

The research objectives can be summarized as a) a remeshing approach to preprocess surface meshes into a representation by semi-regular meshes, which turns out to be more beneficial for CNNs, b) the definition of an autoencoder that handles semi-regular meshes of different size and topology, and by this means c) creating a possibility to transfer and apply trained models to different classes of surface data\footnote{Source code available at: \url{https://github.com/Fraunhofer-SCAI/conv_sr_mesh_autoencoder}}.

Further on in section \ref{sec:rel}, we discuss related work. 
In section \ref{sec:meshes}, we present some theoretical background of surface meshes and relevant characteristics for CNNs, followed by the definition of our convolution and pooling for semi-regular meshes in section \ref{sec:meshcnn}. 
In section \ref{sec:autoencoder}, we introduce our mesh convolutional autoencoder for semi-regular meshes. 
Results for different datasets are presented in section \ref{sec:exp}.

\section{Related Work} 
\label{sec:rel}

\subsection*{Semi-Regular Meshes}

To the best of our knowledge, triangular semi-regular meshes have not been used in the context of convolutional neural networks for graph data of different shape. 
Their piecewise regular structure has been in focus for multi-resolution analysis, because the iterative refinement 
allows analysis at different levels of resolution, which is especially interesting to adapt wavelets to surfaces and geometry compression \cite{Khodakovsky2000,Mallat1989,Payan2015}.
They are generally obtained by remeshing of irregular meshes. In \cite{Khan2020,Payan2015} an overview of semi-regular triangle remeshing algorithms is given.

The authors of \cite{Baque2018} train a neural network on a quadrilateral surface mesh that is mapped to a box. The box's sides coincide with a general grid-based mesh as for images in 2D allowing the application of 2D convolution. Although called semi-regular quadrilateral meshes by the authors, their definition of semi-regular is different than ours, since the regular sides of the boxes are not created by iterative subdivision, which allows us to define the pooling operator. This remeshing approach can only be applied to shapes without boundaries, that can be mapped to a box.
In \cite{Hu2021} a semi-regular mesh structure is used for efficient pooling and unpooling, but the network is not independent of the mesh size and limited to meshes without boundaries.

\subsection*{Convolutional Networks for Graphs and Surfaces}

Generally, there are spectral and spatial convolutional networks for graphs, of which \cite{Bronstein2017,Wu2020} give an overview.

At first, \cite{Bruna2013} exploited the connection of the graph Laplacian and the Fourier basis and they project vertex features to the Laplacian eigenvector basis.
Instead of explicitly computing Laplacian eigenvectors, the authors of \cite{Defferrard2016} use truncated Chebyshev polynomials and in \cite{Kipf2017} they use only first-order Chebyshev polynomials.
These spectral methods require a fixed connectivity of the graph. If not, the basis functions change and the features that the network learns are not guaranteed to be meaningful.

Spatial methods for convolution on graphs aggregate features from the neighbors of the vertices. At first, this idea was presented under the name Neural Network for Graphs \cite{Micheli2009}. Spatial methods allow generalization across different domains and because of their flexibility and efficiency these methods are very popular \cite{Wu2020}. 
Since surface meshes lack a general underlying grid, the orientation of kernels has to be defined with respect to their neighborhood.
To avoid this difficulty, kernels often calculate rotation invariant features, which are sometimes referred to as orientation invariant. 
The authors of \cite{Masci2015,Monti2017} calculate rotation invariant features by averaging the result of different anisotropic kernels that are sensitive to orientation.
In \cite{Boscaini2016a} the kernels are aligned with the principal curvature direction 
and  \cite{Cohen2019, DeHaan2020} introduce anisotropic gauge equivariant kernels that encode orientation information in the features, but are computationally expensive.
The authors of \cite{Cohen2019} demonstrate that for special meshes (in their case the icosahedron having areas with regular connectivity) the anisotropic gauge equivariant convolution can be implemented efficiently.

The work of \cite{Gilmer2017} sums up many spatial approaches by their Message Passing Neural Network. It interprets graph convolutions as a message passing process, in which information is passed from one node to another along the edges.
We want to point out that usually the topology and the number of vertices is fixed, because the pooling \cite{Bouritsas2019,Ranjan2018,Yuan2020} or the spectral convolutional layers require a fixed connectivity \cite{Defferrard2016,Kipf2017,Ranjan2018}. 

It is possible to consider the vertices of meshes describing surfaces as point clouds, which would allow the application of convolutional architectures for 3D point clouds as \cite{Sharp2020, Thomas2019, Wang2019} handling pointwise inputs that consider neighborhoods via kernels. Albeit being the more flexible representation, it lacks an underlying structure that describe the surface, whose deformation we want to analyze \cite{Bouritsas2019}. Also, the mesh provides native connectivity information \cite{Hanocka2019}.

\subsection*{Mesh Convolutional Autoencoders}

The authors of \cite{Tan2018a} present a variational autoencoder for deforming 3D meshes that does not handle meshes but feature representations of the deforming meshes.

In \cite{Ranjan2018} a first convolutional mesh autoencoder (CoMA) has been introduced that handles surface meshes directly. The authors introduced mesh downsampling and mesh upsampling layers, which have a similar effect as pooling and unpooling. They are combined with 
spectral convolutional filters using truncated Chebyshev polynomials as in \cite{Defferrard2016}.
The Neural 3D Morphable Models (Neural3DMM) network presented in \cite{Bouritsas2019} improves those results using the same down and upsampling layers in combination with spiral convolutional layers that break the permutation invariance. The authors of \cite{Yuan2020} apply the CoMA \cite{Ranjan2018} to different datasets and improve the down and upsampling layers slightly. 

The presented mesh convolutional autoencoders work only for meshes of the same size and connectivity, since the downsampling and upsampling layers as well as spectral convolutional layers depend on the adjacency matrix.

The authors of \cite{Hanocka2019} present the MeshCNN architecture that allows an implementation of an encoder and decoder. The edge collapsing based pooling is feature dependent and therefore the low-dimensional mesh embeddings of the deforming meshes can be of different significance. 

\section{Surface Meshes and Their Characteristics}
\label{sec:meshes}

Surfaces in $\mathbb{R}^3$ are generally discretized by triangular polygonal surface meshes.

\begin{mydefi}[Triangular Polygonal Mesh]
A triangular polygonal mesh $\mathcal{M}$ is defined by a set of vertices $V \subset \mathbb{R}^d$ and a set of triangular faces 
$F \subset V \times V \times V$, which describe the shape and point to the vertices they use.
The edges $ E = \{ \{v_1,v_2\} \in V \times V | \exists f \in F \text{ s.t. } v_1 \in f \text{ and } v_2 \in f \} $ are undirected, i.e. if $(v,w) \in E$, $(w,v)$ is also in $E$.
\end{mydefi} 

If the mesh is a surface or manifold mesh, it holds that every edge $e \in E$ is adjacent to at most two faces in $F$. 
A boundary edge $e \in E$ is adjacent to exactly one face in $F$.
The vertices $v \in V$ of a mesh $\mathcal{M}$ store information which is defined by a function $f: V \to \mathbb{R}^m$.
For each vertex $v \in V$, we define the $r$-ring neighborhood $N_r(v)$ as all the vertices $w \in V$ that are connected to $v$ by at most $r$ edges in $E$.
The degree of a vertex $v \in V$ is the size of its one-ring neighborhood $N_1(v)$. 

We refer to a triangular surface mesh as regular, if the degree of all vertices in $V$ is 6 \cite{Payan2015}.
Note, that a regular mesh has limited representation power 
and that not every mesh can be remeshed into a regular one because they 
cannot approximate all types of curvature (see hedgehog theorem \cite{Brouwer1912}).

Images represented in pixels can also be interpreted as two-dimensional surface meshes of rectangular shape in $\mathbb{R}^2$ \cite{Bronstein2017}. 
Every pixel is a vertex, the feature function $f$ outputs the pixels' color values and every vertex is connected by an edge to the eight neighboring pixels (horizontal, vertical and diagonal).

\subsubsection*{Convolution for Images and Surface Meshes}

Although surface meshes in $\mathbb{R}^3$ and images in $\mathbb{R}^2$ both fulfill the definition of polygonal meshes, the overall structure and regularity of the data is highly different, which complicates the application of convolution to surface meshes in $\mathbb{R}^3$ \cite{Bronstein2017}.

CNNs in 2D \cite{Goodfellow2016,LeCun1989} apply the same local filters to local neighborhoods of selected pixels of the image. Because of the global grid structure of the image, the filters can be horizontally and vertically shifted. The filters can be of constant shape and the networks apply them to every local neighborhood. 
As \cite{Cohen2019} pointed out, the shifting of the filters is not well-defined for surface meshes because of the lack of a global grid. Also, the local neighborhoods of a surface mesh can have any size and arrangement as long as they are locally Euclidean.

For images, the size of each instance is usually constant. Since the samples are of rectangular shape, they can easily be resized or padded into the desired size.
The constant size allows one neural network to handle all the data and allows the use of existing networks as pretrained prototypes for different applications. 
However, the size of surface meshes varies strongly in general.  
Table \ref{tab_conv} gives on overview of the mentioned mesh characteristics of images and surface meshes. 
The authors of \cite{Bronstein2017} summarize that both the similar structure for local neighborhoods and the underlying global grid are reasons why CNNs work so efficiently for images.

As mentioned, one cannot enforce a regular mesh discretization for every surface in $\R^3$, which would lead to an underlying global grid \cite{Brouwer1912}. We aim to enforce a similar structure in the local neighborhoods by choosing a different approximation of the surface. In this way, an efficient application of convolution on surface meshes becomes possible. Note, that remeshing the polygonal mesh only changes the representation of the objects. The considered surface embedded in $\R^3$ is the same, but now represented by a different discrete approximation.

\bgroup
\newcommand\cp{0.35}
\def\arraystretch{1}
\begin{table}
\begin{center}
\begin{tabular}{| @{\hspace{\cp em}} c @{\hspace{\cp em}} | @{\hspace{\cp em}} c @{\hspace{\cp em}} | @{\hspace{\cp em}} c @{\hspace{\cp em}} | @{\hspace{\cp em}} c @{\hspace{\cp em}} |}
\hline
  & \multirow{2}{*}{2D Image} & Surface &  Semi-Regular \\
  &  & Mesh & Surface Mesh \\ 
\hline\hline
 data & \multicolumn{3}{c|}{information saved on vertices} \\ \hline
 grid      & global  & locally & locally \\
 structure & structure & Euclidean & Euclidean \\\hline
 connectivity & fixed  & - & semi-regular \\ \hline
 distance to  & \multirow{2}{*}{fixed} & \multirow{2}{*}{-} & \multirow{2}{*}{-} \\
 neighbors    & & & \\ \hline
 
\multirow{3}{*}{\shortstack{size of \\ instances} } & \multirow{3}{*}{similar} & \multirow{3}{*}{\shortstack{highly \\ different}} & highly different  \\
  & & & with similar  \\
  & & & local patches \\
\hline
\end{tabular}
\end{center}
\caption{Characteristics of images and surface meshes in $\R^3$ relevant for CNNs}
\label{tab_conv}
\end{table}
\egroup

\subsubsection*{Semi-Regular Meshes}

Semi-regular meshes are a flexible representation of surfaces in $\R^3$, that have a regular local structure and allow irregularities at selected vertices, whose positions we can control \cite{Payan2015}. The overview in Table \ref{tab_conv} shows that the mesh characteristics of semi-regular meshes are closer to the ones of 2D images than compared to general surface meshes.

\begin{figure*}[t]
\begin{center}
\begin{minipage}{\linewidth}
  \centering
  \raisebox{-0.5\height}{{\includegraphics[width=.3\linewidth, trim=8.6cm 8.3cm 4cm 8.2cm, clip]{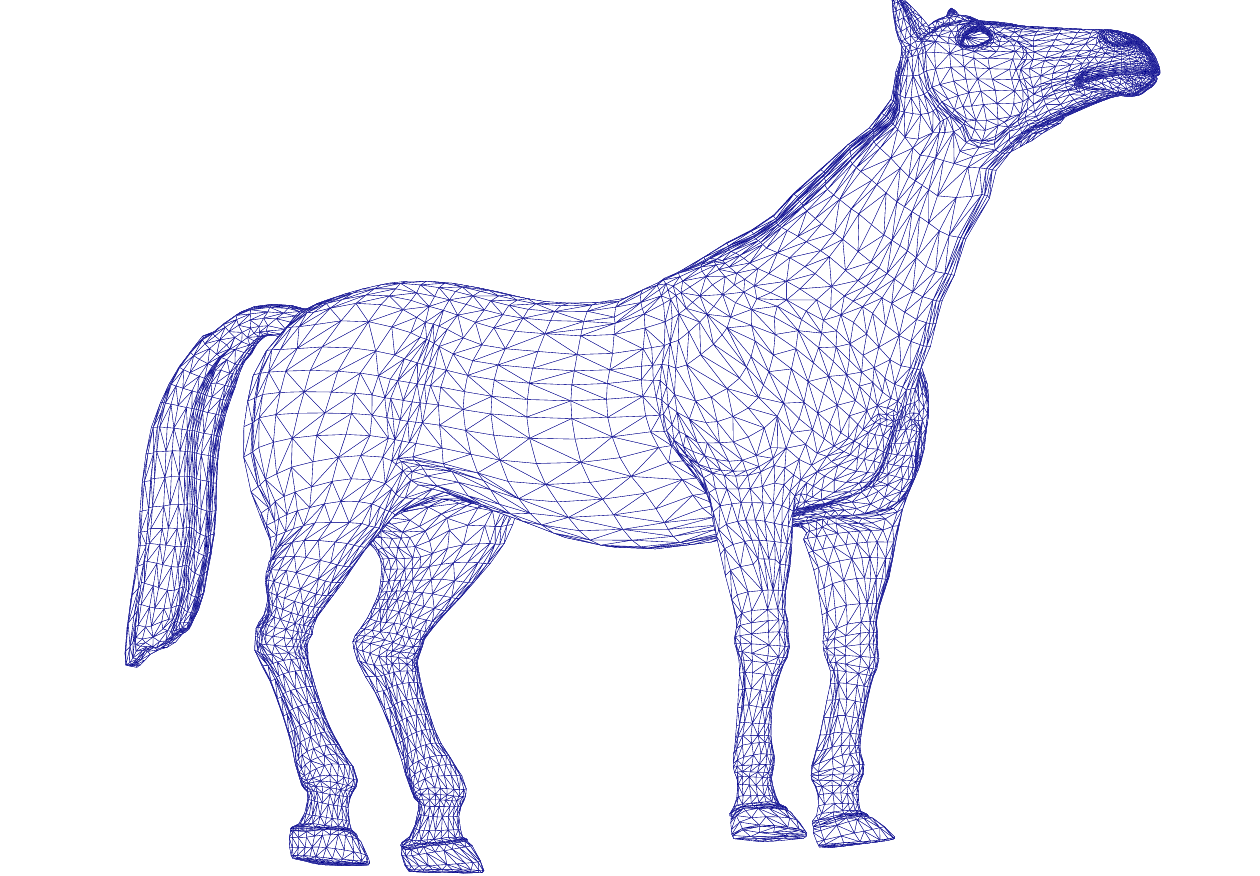}}}  
  \raisebox{-0.5\height}{\Large$\Rightarrow$}
  \raisebox{-0.5\height}{{\includegraphics[width=.3\linewidth, trim=8.6cm 8.3cm 4cm 8.1cm, clip]{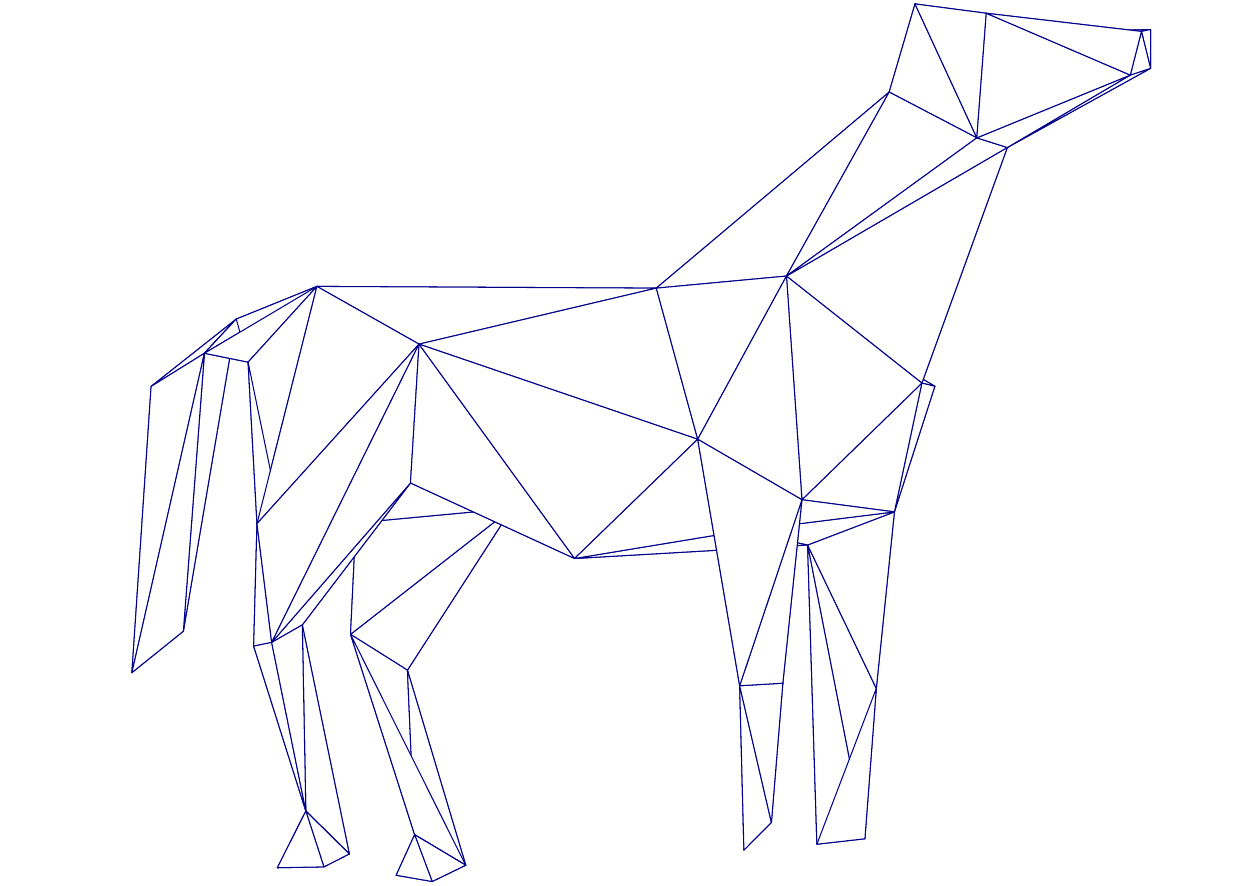}}}  
  \raisebox{-0.5\height}{\Large$\Rightarrow$}
  \raisebox{-0.5\height}{{\includegraphics[width=.3\linewidth, trim=8.6cm 8.3cm 4cm 8.2cm, clip]{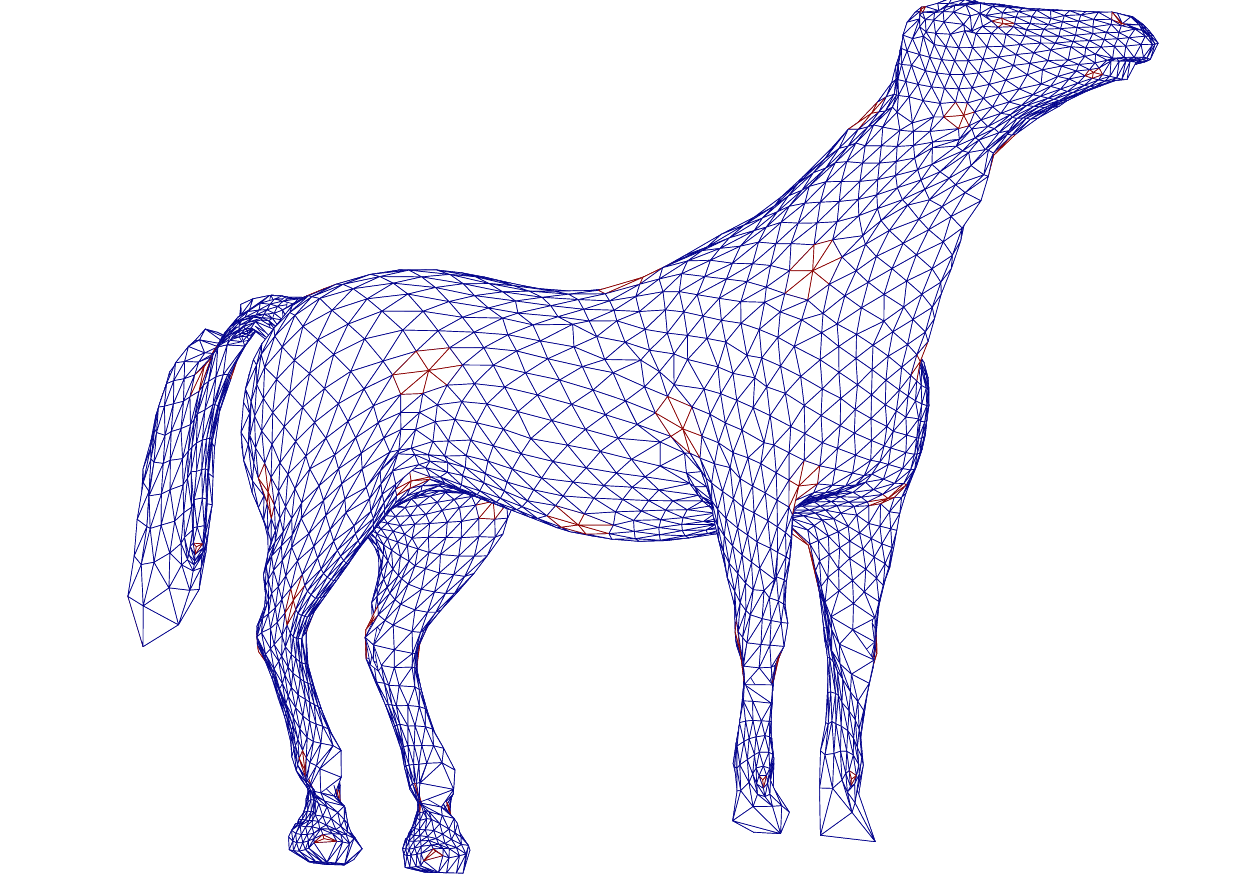}}}  
\end{minipage}
\begin{minipage}{.98\linewidth}
 \begin{tabular}{>{\centering\arraybackslash}p{0.3\linewidth} >{\centering\arraybackslash}p{0.3\linewidth} >{\centering\arraybackslash}p{0.3\linewidth}}
    \textbf{Irregular surface mesh}  & \textbf{Low resolution base mesh} & \textbf{Semi-regular mesh} \\
 \end{tabular}
\end{minipage}
\end{center}
  \caption{Remeshing of the horse template mesh. After coarsening the original mesh to a low resolution base mesh, the faces are subdivided three times. In the semi-regular mesh the faces adjacent to an irregular vertex are highlighted in red.}
\label{semiregular}
\end{figure*}

We follow the definition of semi-regular meshes from \cite{Payan2015}, which gives one condition on the specific structure: iteratively merging four triangular faces of a semi-regular mesh into one leads to a low-resolution mesh. This means that all vertices are regular (i.e.\ have six neighbors) besides the vertices of the low-resolution mesh (see Figure \ref{semiregular}).
Therefore, a semi-regular mesh is obtained by regular subdivision of a low resolution mesh that can be irregular.
We refer to the faces of the low-resolution mesh that are iteratively subdivided as regional patches.

Note that the iterative subdivision of the low-resolution mesh automatically defines a multi-scale structure. This is why semi-regular meshes are well suited for multi-resolution analysis \cite{Lee1998,Payan2015}. Later on, this structure allows us to define a local pooling operator on the semi-regular meshes.

\subsection*{Remeshing} 

At first, a coarse approximation of the input mesh is built. To coarsen the surface meshes we employ a slightly adapted Garland-Heckbert-algorithm for surface simplification using quadric error metrics \cite{Garland1997}.

Afterwards, every face of the coarse base mesh is iteratively subdivided into four new faces for a given number of times. All newly created vertices have six neighbors. 
When the desired level of refinement is reached, the resulting semi-regular mesh geometry has to be fit to the original irregular mesh in order to describe the surface well.
If for one deforming shape the mesh topology is constant over time, it is enough to remesh a template mesh. The semi-regular remeshing result can be transferred to meshes of the same topology.

We provide more details to the remeshing algorithm in the supplementary material.

\section{Convolution and Pooling for Semi-Regular Meshes} 
\label{sec:meshcnn}

Convolutional kernels consider local features in the neighborhood of the vertices. 
The size of the considered neighborhood $N_r(v)$ is given by the kernel size $r$. 
In a convolutional layer, a set of kernels is applied to every vertex of the input. 
Generally, the number of vertices and the mesh's connectivity have to be constant. Only the vertex features change. 

Since the convolutional networks learn local features, we propose to input every regional patch of the semi-regular mesh separately. This allows us to handle meshes of different size.
To the best of our knowledge, we thereby present the first mesh convolutional autoencoder that handles meshes of different size.
In order to not lose the embedding of the regional patch in the whole mesh, we consider the environment of the patches via padding.

\subsubsection*{Hexagonal Convolution}

The regional patches are of the same regular structure. All vertices have exactly six neighbors, only the three corners can be irregular, but we project their neighborhood to a regular one.
Additionally, the patches are intrinsically two-dimensional and represent a surface. 
Therefore, the application of a 2D-convolutional kernel is possible. Since the regional patches are represented in hexagonal grids, the application of hexagonal 2D-convolutional kernels has shown to give better results \cite{Hoogeboom2018,Steppa2019}. Similarly to \cite{Baque2018}, the consistent degree of the vertices results in better runtimes since similar calculations at these vertices can be moved to GPU.

On the local regular structure, the translation of the convolutional kernels is well-defined. Therefore the kernels preserve the orientation of the neighborhood and are anisotropic. The padded patch based approach assures gauge equivariance of the network.
The authors of \cite{Cohen2019,DeHaan2020} show how anisotropic kernels that preserve orientation significantly improve the expressivity of models.

Note that the network does not correct differences in the distances to neighbors or angles between neighbors. The edge lengths of the semi-regular meshes are stable, because of the edge length regularization during the remeshing.

\begin{figure*}[t]
\begin{center}
\begin{minipage}{0.96\linewidth}
  \centering
  \raisebox{-0.5\height}{{\includegraphics[width=.15\linewidth, trim=1.5cm 2.3cm 2.6cm 0.7cm, clip]{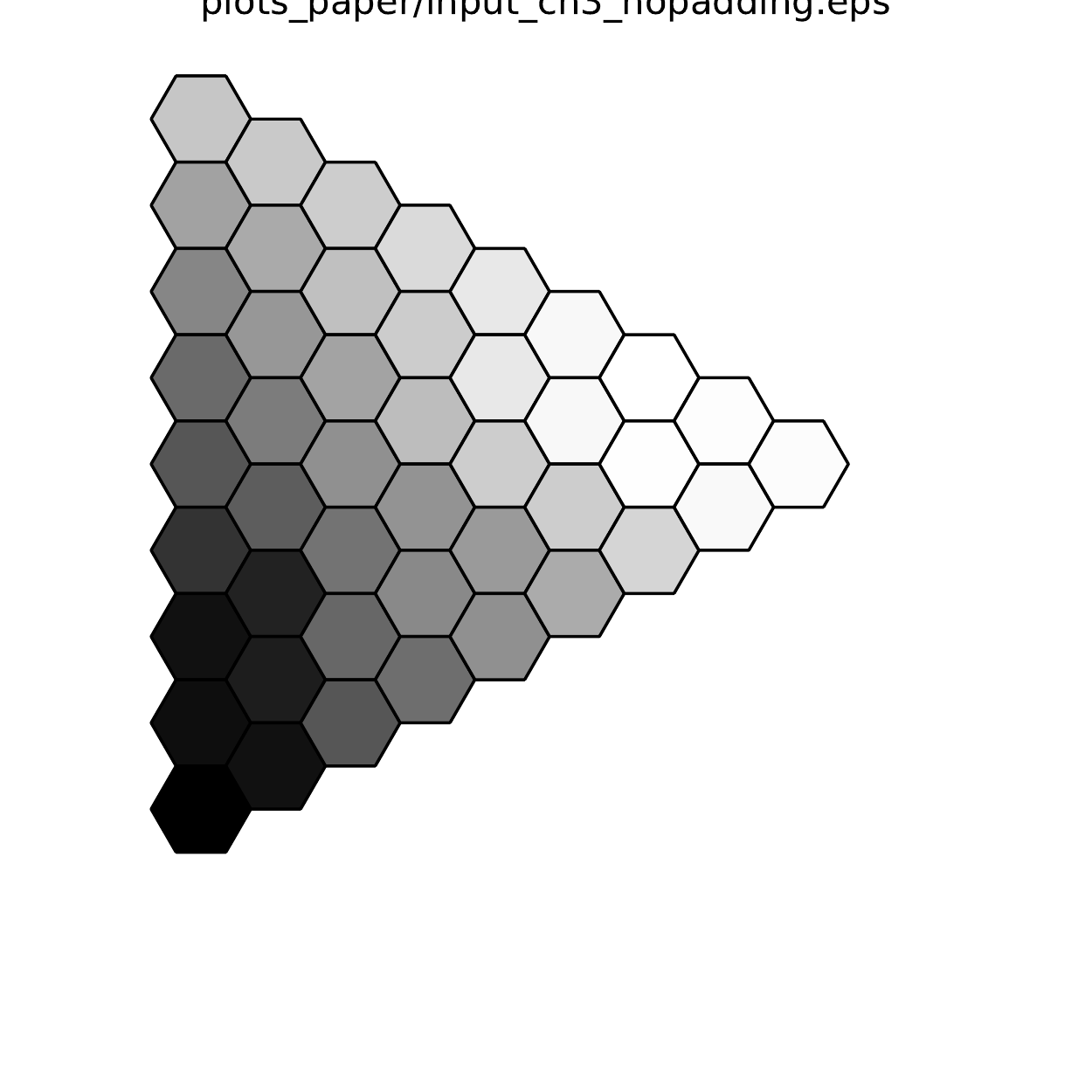}}}  \raisebox{-0.5\height}{\Large$\Rightarrow$}
  \raisebox{-0.5\height}{{\includegraphics[width=.19\linewidth, trim=1.5cm 2.3cm 2.6cm 0.7cm, clip]{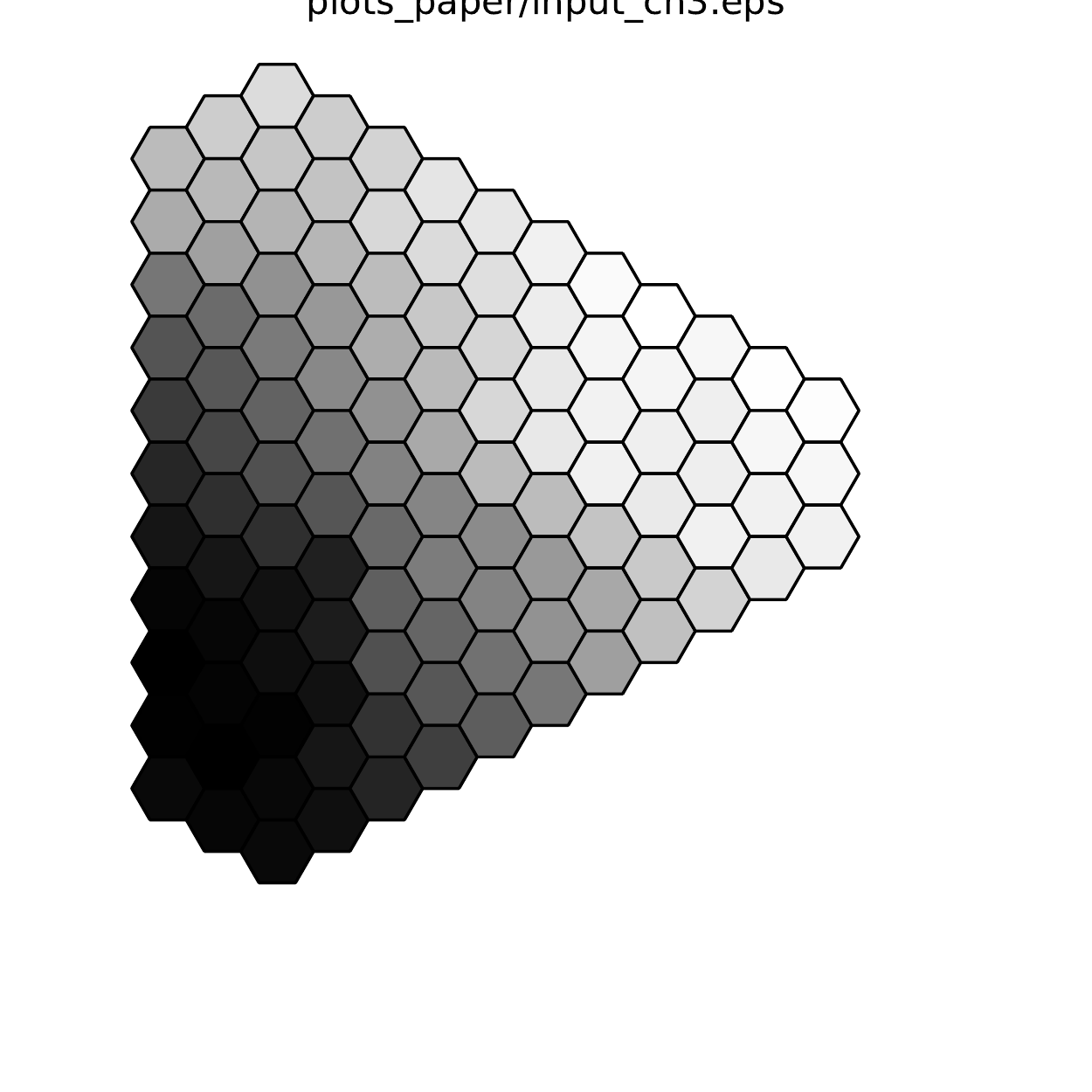}}}  \raisebox{-0.5\height}{\Large$\Rightarrow$}
  \raisebox{-0.5\height}{{\includegraphics[width=.08\linewidth, trim=1.9cm 2.3cm 2.8cm 1cm, clip]{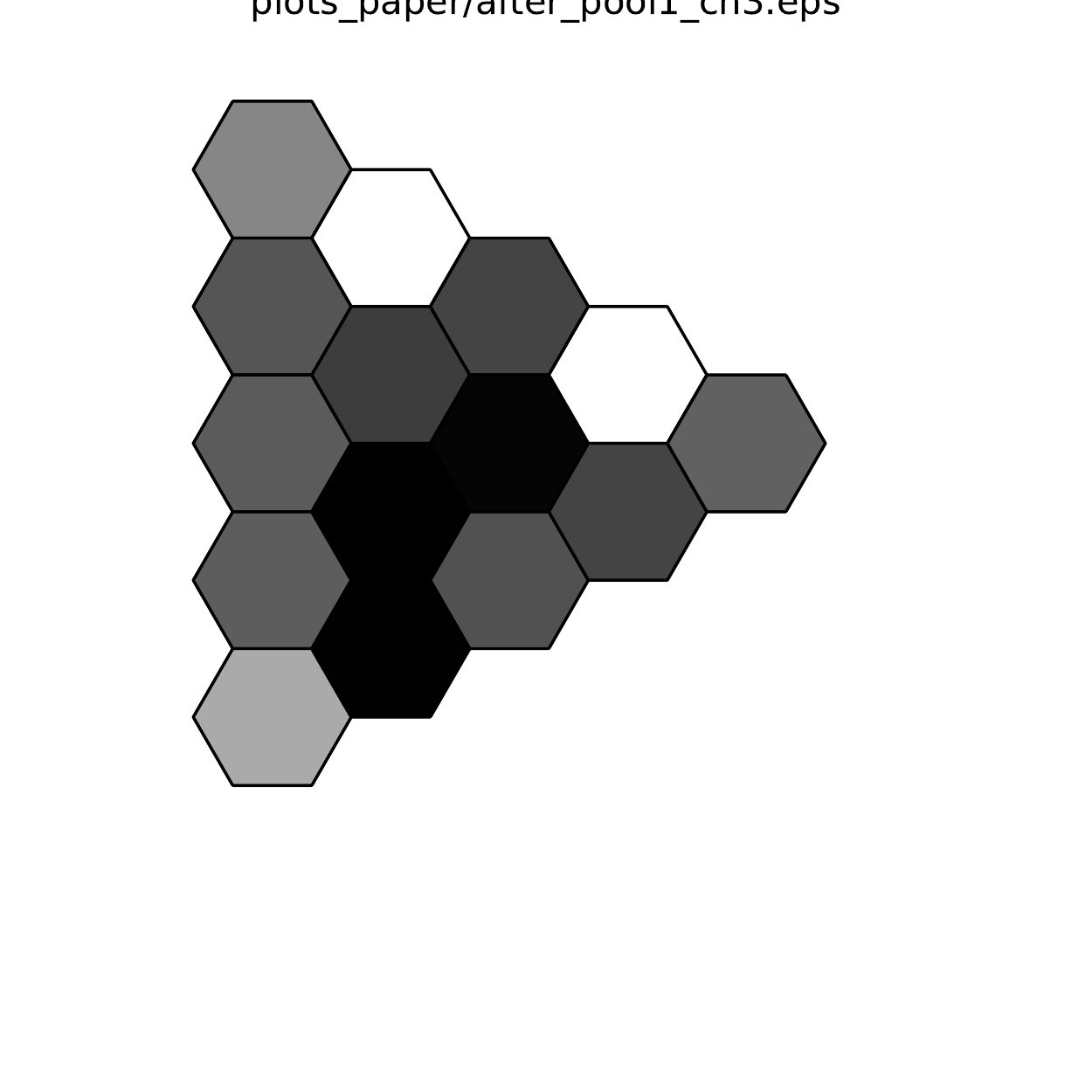}}} 
  \raisebox{-0.5\height}{\Large$\Rightarrow$}
  \raisebox{-0.5\height}{{\includegraphics[width=.05\linewidth, trim=2.5cm 4cm 3.1cm 1.2cm, clip]{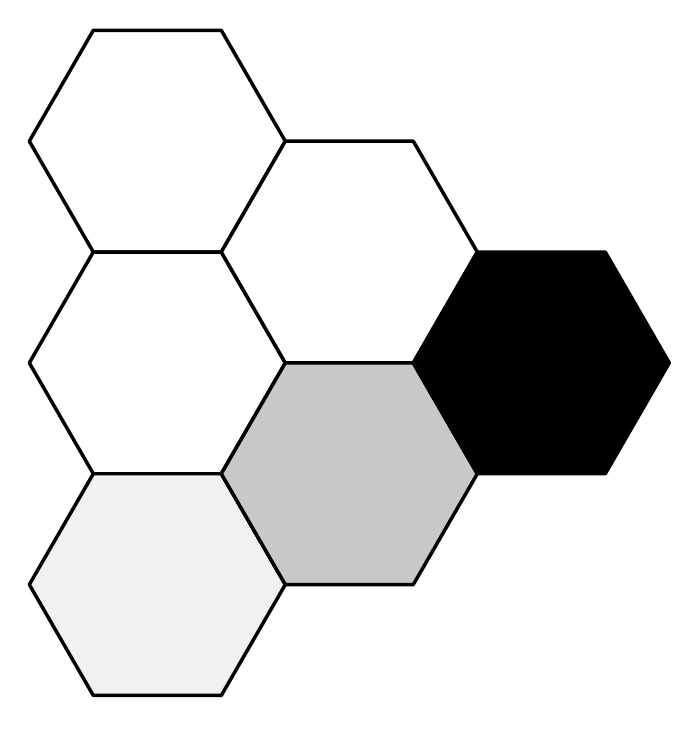}}} 
  \raisebox{-0.5\height}{\Large$\Rightarrow$}
  \raisebox{-0.5\height}{{\includegraphics[width=.015\linewidth, trim=7.4cm 2.8cm 7cm 1.2cm, clip]{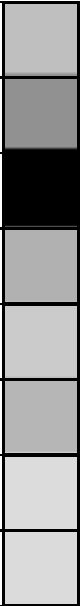}}} 
  \raisebox{-0.5\height}{\Large$\Rightarrow$}
  \raisebox{-0.5\height}{{\includegraphics[width=.08\linewidth, trim=1.9cm 2.3cm 2.8cm 1cm, clip]{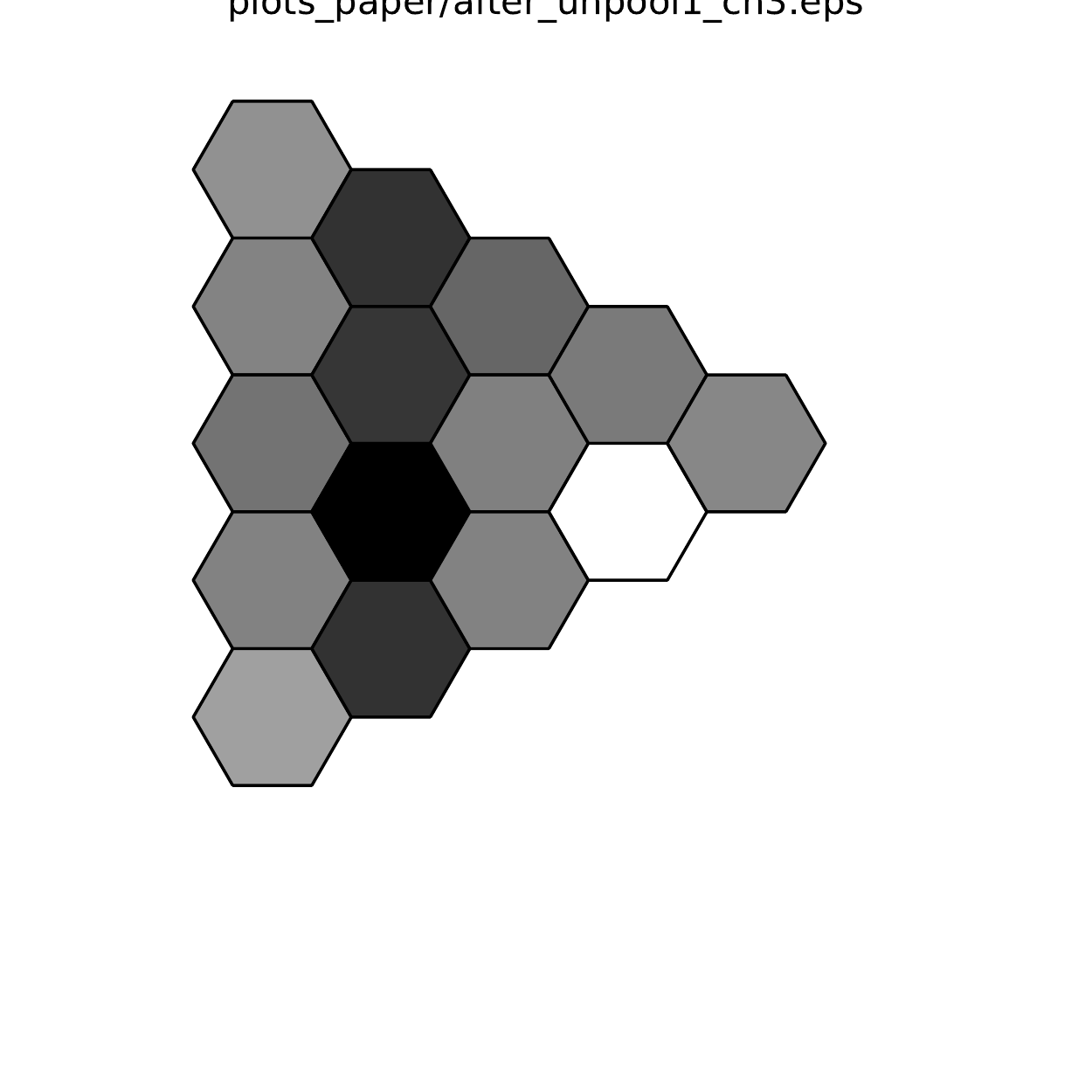}}} 
  \raisebox{-0.5\height}{\Large$\Rightarrow$}
  \raisebox{-0.5\height}{{\includegraphics[width=.15\linewidth, trim=1.5cm 2.3cm 2.6cm 0.7cm, clip]{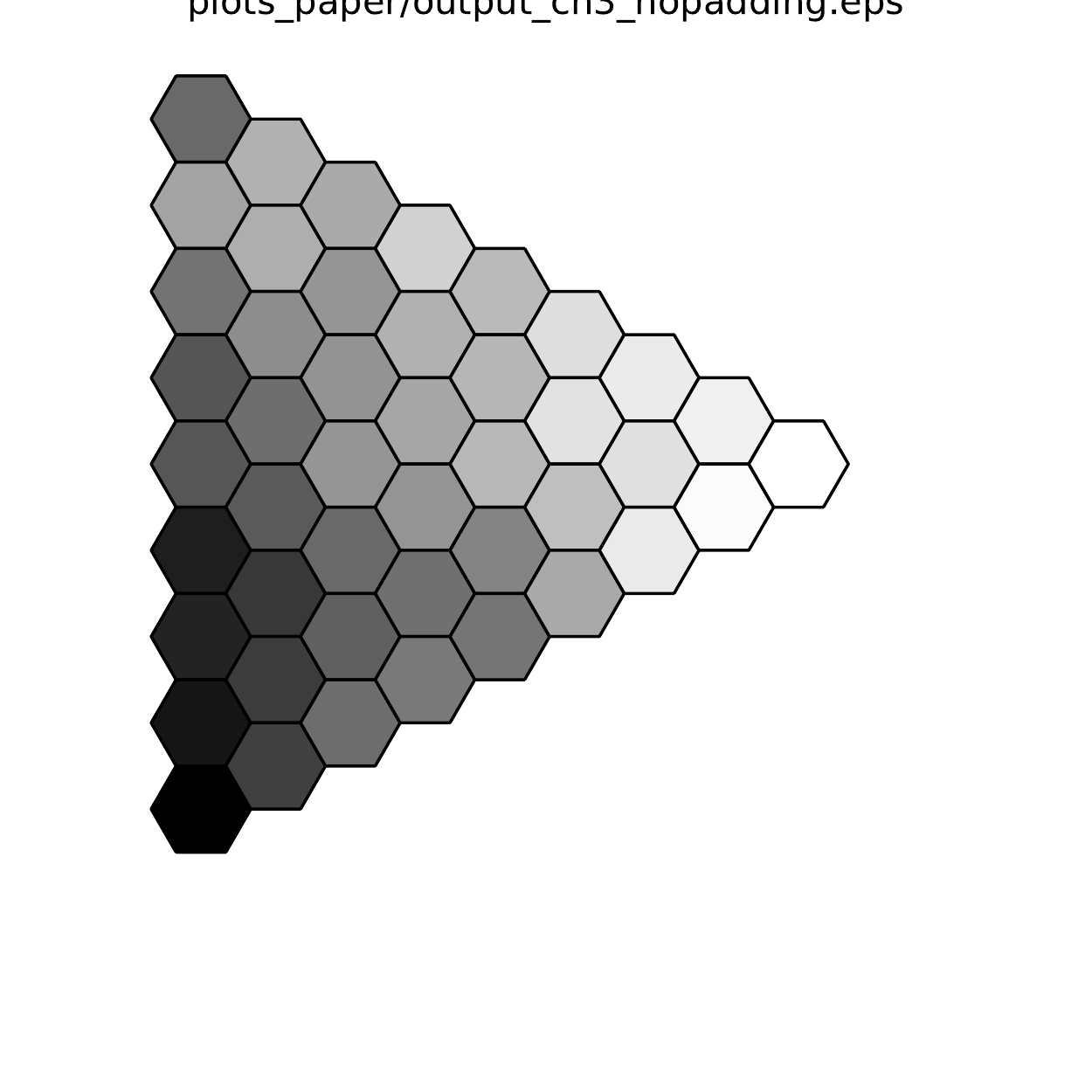}}}
\end{minipage}
\begin{minipage}{0.97\linewidth}
 \begin{tabular}{p{0.18\linewidth} p{0.19\linewidth} p{0.16\linewidth} >{\centering\arraybackslash}p{0.07\linewidth} p{0.1\linewidth} p{0.12\linewidth}}
    \textbf{Input patch} & \textbf{Padded Input patch}  & \textbf{Pooling} & \textbf{Embed-ding} &\textbf{Unpooling}  & \textbf{Output Patch} \\
 \end{tabular}
\end{minipage}
\end{center}
  \caption{The pooling layers change the resolution of the semi-regular meshes. We input the padded regular patch and apply pooling twice, which undoes the subdivision of the faces. In the decoder the unpooling increases the resolution again by subdividing the faces.}
\label{pooling}
\end{figure*}

\subsubsection*{Pooling}

The piecewise regular form of the semi-regular meshes has a multi-scale structure, which is created by the iterative subdivision of the faces of the low-resolution mesh.
We take advantage of this structure that all semi-regular meshes have in common, and define an average pooling operator, that undoes the subdivision of one into four faces. Herewith, we reduce the dimensions of the features and the number of network parameters, as pooling layers for 2D convolution do as well \cite{Goodfellow2016}.
Since we apply average pooling, the vertices that are kept during the pooling take the average of their own value and the values of the neighboring vertices in the one-ring neighborhood that are removed.

To increase the resolution of the mesh patches in the decoder, the unpooling operator recreates the multi-scale structure of the semi-regular mesh. Every face is subdivided into four faces. The newly created vertices are assigned the average value of neighboring vertices from the lower-resolution mesh patch.
Figure \ref{pooling} illustrates how the pooling and unpooling layers undo the subdivision of the regular patch or increase its resolution respectively.

\subsubsection*{Padding}

The padding is crucial for the network to consider the regional patches in a larger context. Since the network handles the patches separately, we consider the features of the neighboring patches in the padding.

We apply a padding of the size of the first layer's kernel size. The padding considers the vertices of the neighboring patches. 
If the number of neighbors is lower than six, we interpolate the values for empty vertices. If the number is higher than six, we take the closest vertices in both cyclic rotations. If the vertices are boundary vertices of the whole mesh, we decide to pad the patch with the boundary vertices' features. 
Figure \ref{pooling} shows a padded regular patch.

\section{Autoencoder for Semi-Regular Meshes of Different Size} 
\label{sec:autoencoder}

The network handles the regional patches separately, which 
allows us to handle meshes of different sizes.  The information of the neighboring patches is not lost but included in the padding.

The mesh autoencoder consists of an encoder and a decoder. The encoder consists of 2 hexagonal convolutional layers (implementation of \cite{Steppa2019}) 
of kernel sizes 2 and 1. Each of the convolutions is followed by a biased ReLU \cite{Glorot2011}. The average pooling layers (see section \ref{sec:meshcnn}) 
are interleaved between convolutional layers. 
The encoder transforms every padded patch, which corresponds to one face of the low-resolution mesh, from $\R^{111\times3}$ to an 8-dimensional latent vector using a fully connected layer at the end. 

Following the decoder’s fully connected layer, 2 hexagonal convolutional layers (followed by a biased ReLU) with interleaved average unpooling layers reconstruct the patches. Each unpooling layer subdivides every face into 4 faces, following the subdivision process of the remeshing. The last layer is a hexagonal convolutional layer without activation function that reduces the number of features to three dimensions.
A detailed structure of the network is given in Table \ref{network} supplied as additional material together with the distribution of the 18184 trainable weights.  

Note that we are able to handle non-manifold edges of the coarse base mesh because the patches, whose interiors by construction have only manifold-edges, are fed separately.
Figure \ref{pooling} illustrates the patch sizes inside the encoder and decoder.

\section{Experiments}
\label{sec:exp}
We test our convolutional autoencoder for semi-regular meshes on four different datasets and compare the achieved reconstruction errors to state-of-the-art models.

\subsection*{Datasets}
\textbf{GALLOP:} 
The authors of \cite{Sumner2004} present a dataset containing triangular meshes representing a motion sequence from a galloping horse, elephant, and camel. Each sequence has 48 timesteps.
The three animals move in a similar way but the meshes that represent the surfaces of the three animals are highly different in connectivity and in the number of vertices (horse: 8,431, camel: 21,887, elephant: 42,321). This is why the authors of \cite{Yuan2020} trained three different mesh autoencoders as presented in \cite{Ranjan2018}.
The surface approximations are remeshed to semi-regular meshes for each animal. The coarse base meshes of approximately 110 faces are subdivided 3 times. The new meshes are still of different connectivity, but all are made up of regional regular patches. The resulting numbers of vertices are listed in Table \ref{errors}. 

We normalize the 3D coordinates to $[-1,1]$ relative to the coordinates' ratio and translate every patch to zero mean. For the training of CoMA \cite{Ranjan2018} and Neural3DMM \cite{Bouritsas2019} every vertex was normalized to zero mean and a standard deviation of one. The patch based approach does not allow a vertex wise normalization if we want to learn and transform the local deformation. This is why for the baseline training of the CoMA and Neural3DMM on the remeshed data, we normalize the whole mesh to zero mean and standard deviation of one.

We use the first 75\% of the galloping sequence of the horse and camel for training of the network. The architecture is tested on the remaining 25\% and the whole sequence of the elephant, which is never seen during the training.

\textbf{FAUST:} 
We select 100 meshes from the FAUST dataset \cite{Bogo2014}, which are in correspondence to each other, to be able to compare to the other architectures.
The dataset consists of 10 different bodies in 10 different poses. The irregular surface meshes have 6890 vertices. We conduct two different experiments: at first we consider known poses of two unseen bodies in the testing set. Then we consider two unknown poses of all bodies in the testing set. In both cases, 20\% of the data is included in the testing set.

The meshes are remeshed to a semi-regular mesh representation. The data and the mesh patches are normalized in the same way as for the GALLOP dataset.

\textbf{TRUCK and YARIS:} 
In a car crash simulation the car components, which are generally represented by surface meshes, often deform in different patterns. Every component is discretized by a surface mesh, while the local deformation is described by the same physical rules.

The dataset TRUCK consists of 32 completed frontal crash simulations of a Chevrolet C2500 pick-up truck (from NCAC \cite{NCAC}), using the same truck, but with different material characteristics, which is a similar setup to \cite{Bohn2013}.  For this setup the authors of \cite{Hahner2020} detect patterns in the deformation of the components using a general representation by oriented bounding boxes, which allowed them to train one autoencoder for the whole car model. 
For our analysis, we select 6 components (the front and side beams), whose meshes are remeshed to semi-regular meshes, and 30 equally distributed time steps. The model is trained on 30 simulations and 70\% of the timesteps. The trained architecture is tested on 2 complete simulations and on the remaining timesteps for the other 30 simulations.

To study the transfer learning capacities of our architecture, we test the architecture that is trained on the TRUCK dataset on a different dataset, which also contains deforming components from  a different car crash simulation.
The YARIS dataset consists of 10 completed frontal crash simulations of a detailed model of the Toyota Yaris (from NCAC \cite{NCAC}) with different material characteristics. 
For our analysis, we select 10 components (the front and side beams plus the crashbox), whose meshes are remeshed to semi-regular meshes, and 26 equally distributed time steps. Figures \ref{yarisparts} and  \ref{truckparts} in the supplementary material visualize the selected components for the car crash simulations. This data will be made available after acceptance of the article. 

We normalize the meshes that discretize car components to zero mean and range $[-1,1]$ relative to the coordinates' ratio. Every patch is translated to zero mean. 

\begin{figure}
\begin{center}
\begin{minipage}{0.85\linewidth}
  \centering
  \raisebox{-0.5\height}{{\includegraphics[width=.28\linewidth, trim=2cm 3.4cm 2.9cm 1cm, clip]{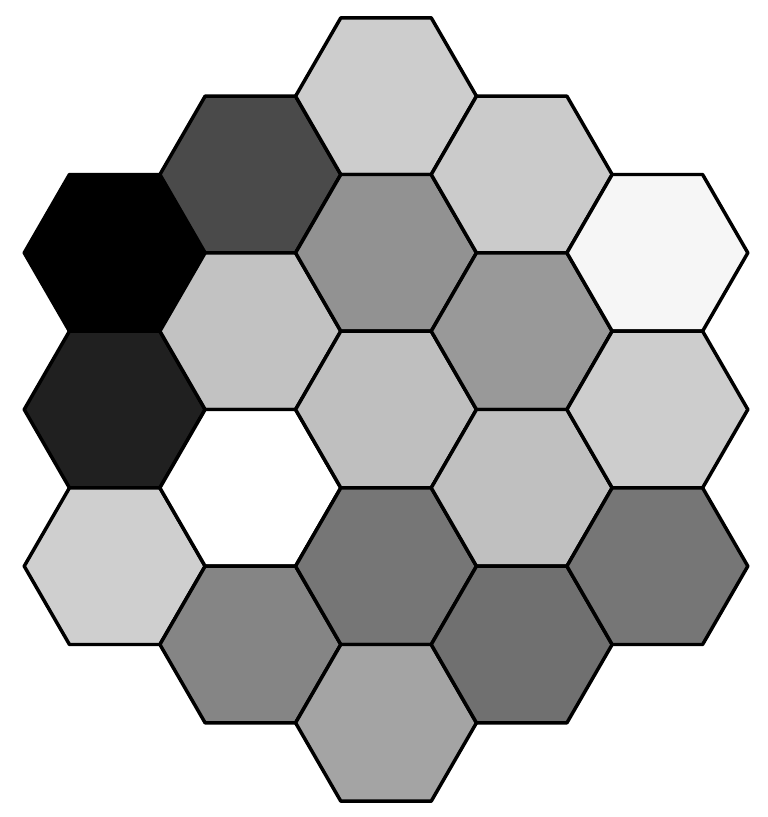}}}
  \raisebox{-0.5\height}{{\includegraphics[width=.28\linewidth, trim=2cm 3.4cm 2.9cm 1cm, clip]{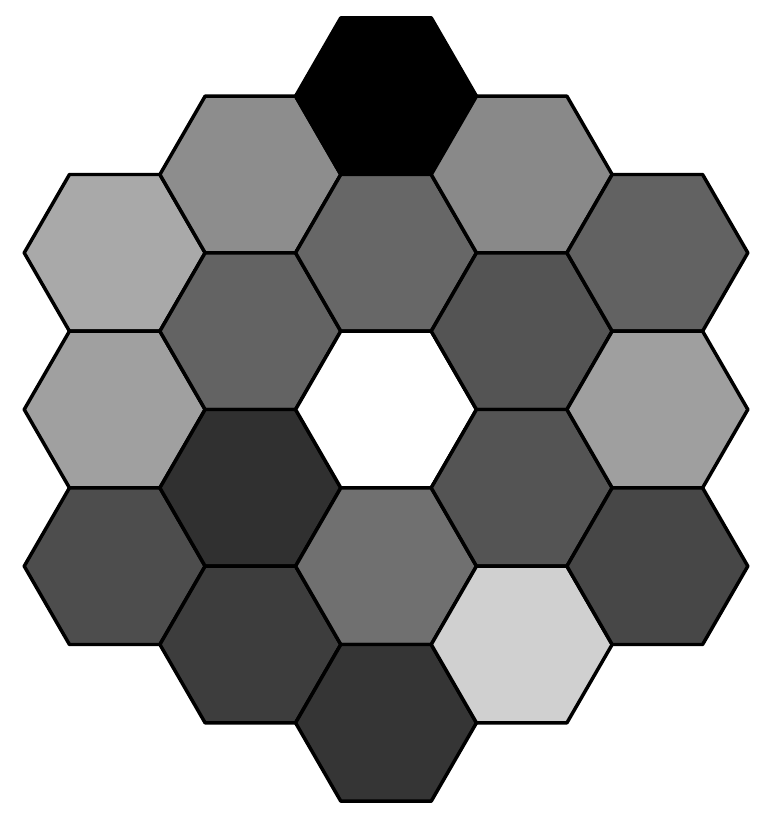}}}
  \hspace*{.01in}
  \raisebox{-0.5\height}{{\includegraphics[width=.18\linewidth, trim=2.5cm 3cm 3.2cm 2.3cm, clip]{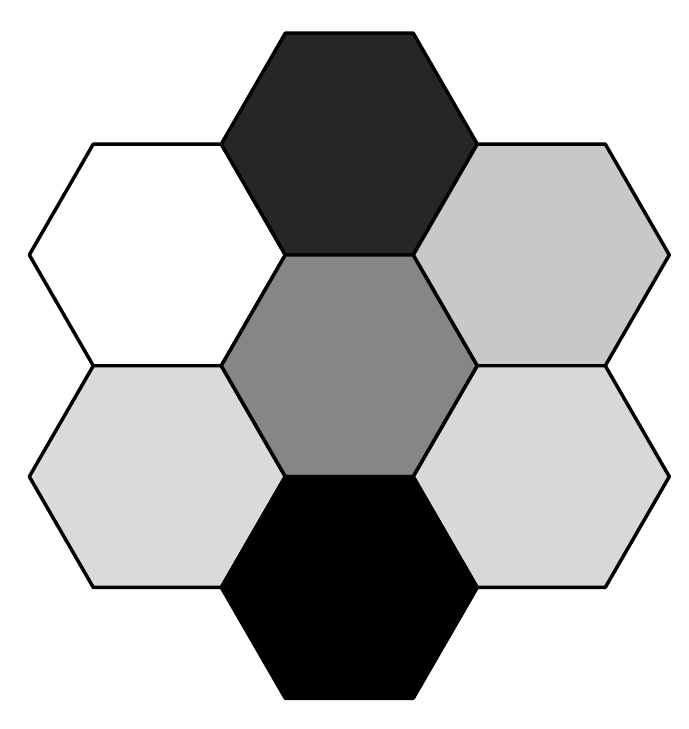}}}
  \raisebox{-0.5\height}{{\includegraphics[width=.18\linewidth, trim=2.5cm 3cm 3.2cm 2.3cm, clip]{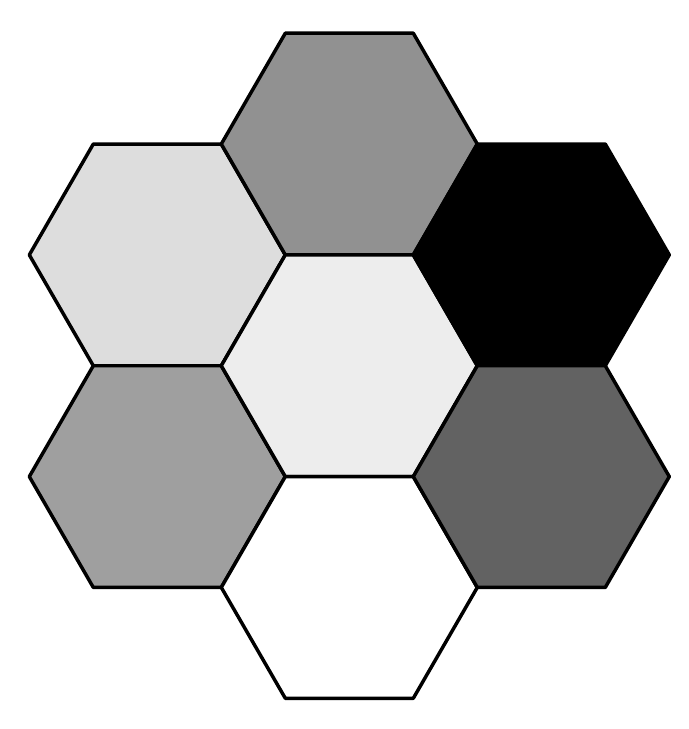}}}
\end{minipage}
\end{center}
\caption{Anisotropic filters from the first and second 
convolutional layers of the encoder trained on the GALLOP dataset.}
\label{anisotropickernel}
\end{figure}

\begin{figure*}

\renewcommand{\arraystretch}{0}
\setlength{\tabcolsep}{0em}
\begin{center}
\begin{minipage}{\linewidth}
  \centering
 \begin{tabular}{p{0.07\linewidth} 
 >{\centering\arraybackslash}p{0.22\linewidth}  
 >{\centering\arraybackslash}p{0.237\linewidth}
 >{\centering\arraybackslash}p{0.237\linewidth}
 >{\centering\arraybackslash}p{0.237\linewidth}}
 
  &\textbf{Original Mesh} & \textbf{CoMA \cite{Ranjan2018} } & \textbf{Neural3DMM \cite{Bouritsas2019} }& \textbf{Our Reconstruction} \\

  Camel $t=39$ & \scalebox{-1}[1]{\raisebox{-0.5\height}{{\includegraphics[width=\linewidth, trim=11cm 14.5cm 26.7cm 17cm, clip]{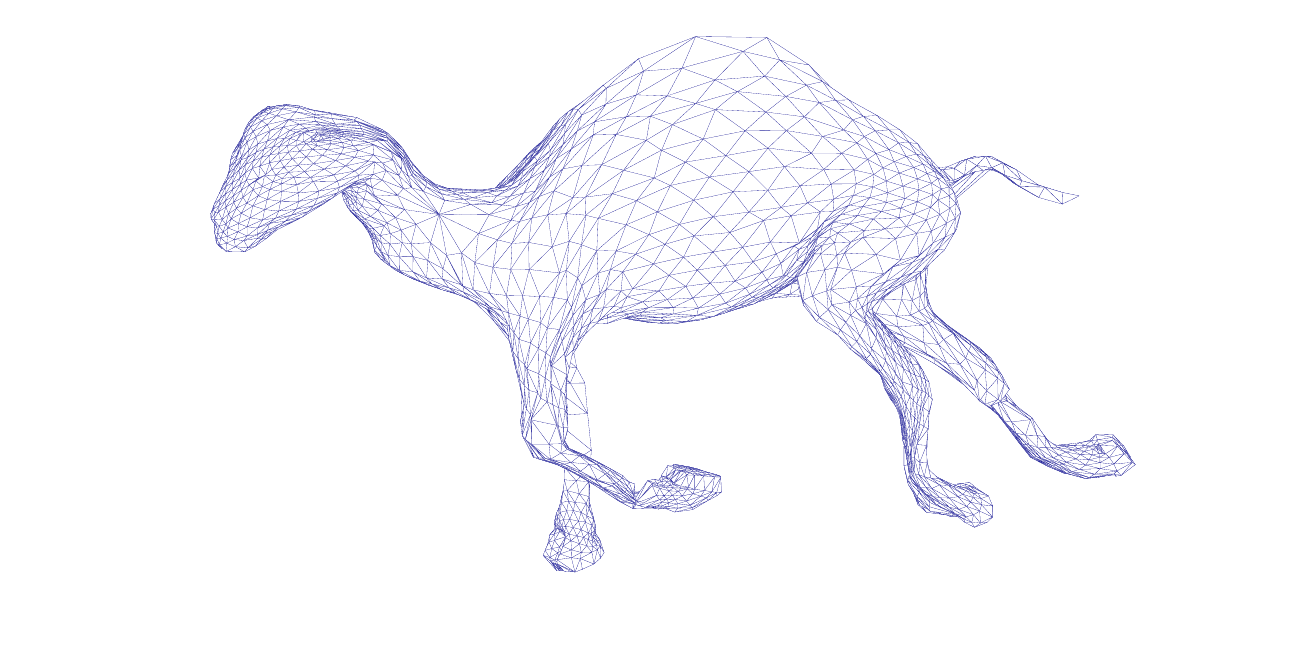}}}} &
  \scalebox{-1}[1]{\raisebox{-0.5\height}{{\includegraphics[width=\linewidth, trim=11cm 14.5cm 27cm 17.5cm, clip]{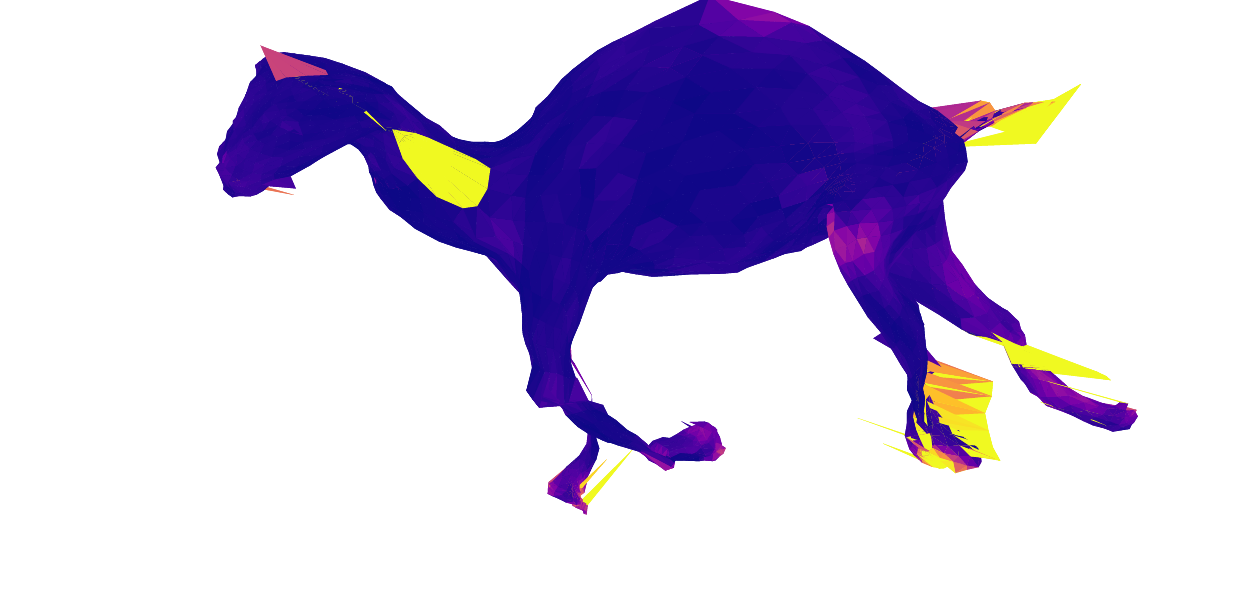}}}} &
  \scalebox{-1}[1]{\raisebox{-0.5\height}{{\includegraphics[width=\linewidth, trim=11cm 14.5cm 27cm 17.5cm, clip]{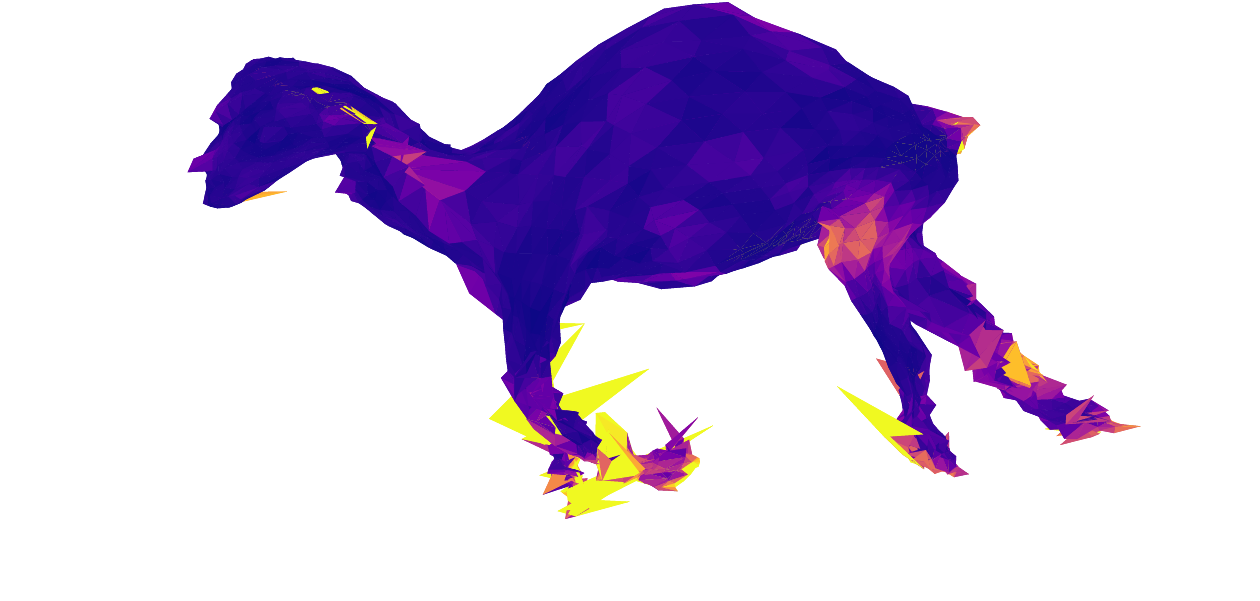}}}}  &
  \scalebox{-1}[1]{\raisebox{-0.5\height}{{\includegraphics[width=\linewidth, trim=11cm 14.5cm 27cm 17cm, clip]{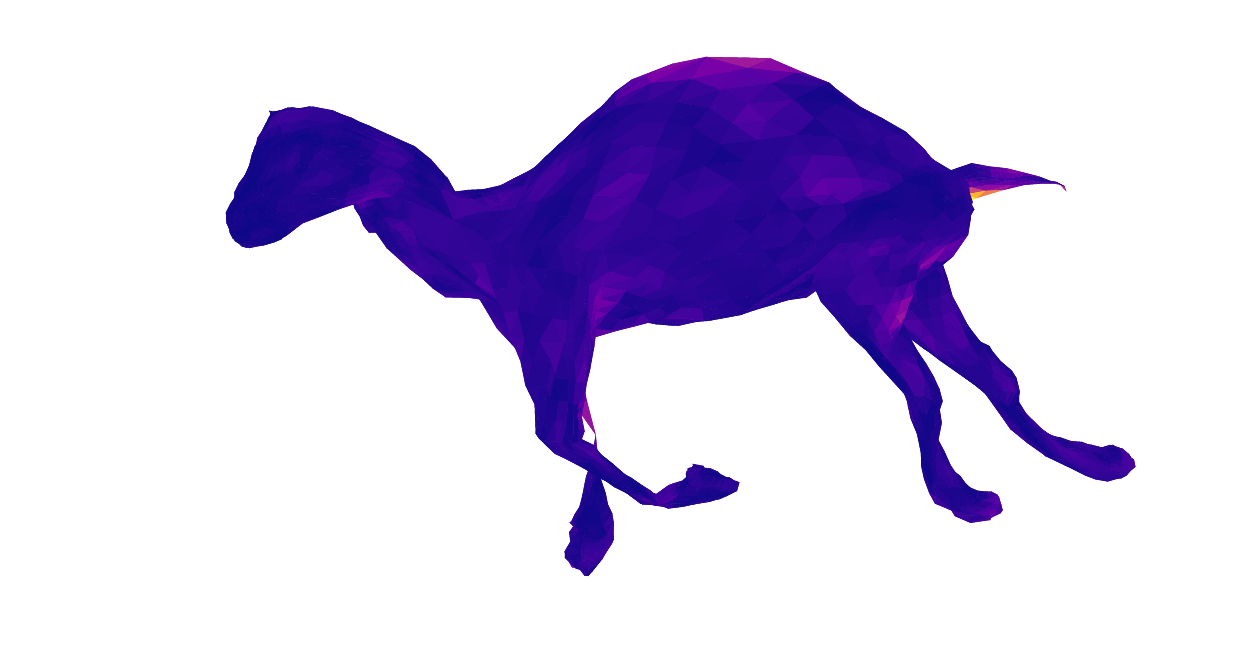}}}} \\
  FAUST \mbox{unknown} pose & 
  \raisebox{-0.5\height}{{\includegraphics[width=0.8\linewidth, trim=13cm 12.5cm 28.5cm 14cm, clip]{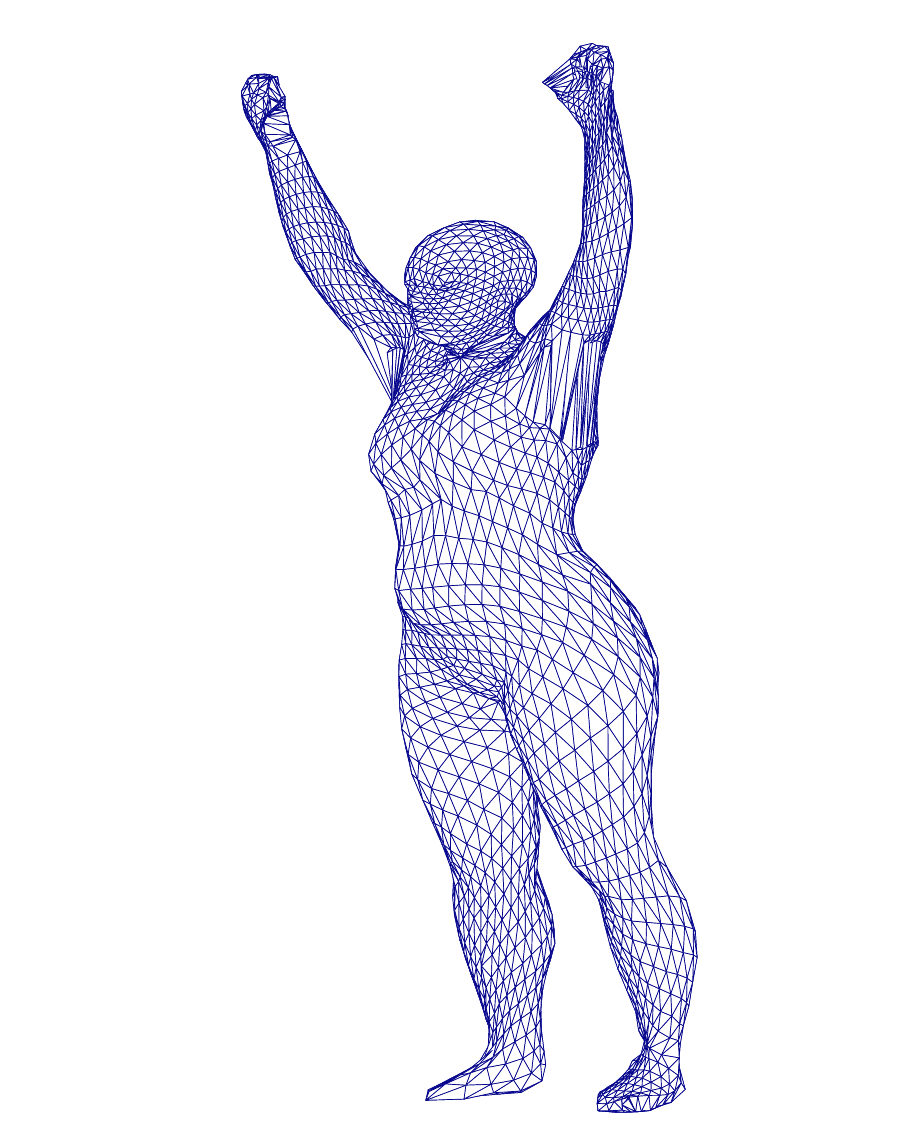}}} &  
  \raisebox{-0.5\height}{{\includegraphics[width=0.9\linewidth, trim=12.5cm 12.2cm 28cm 15cm, clip]{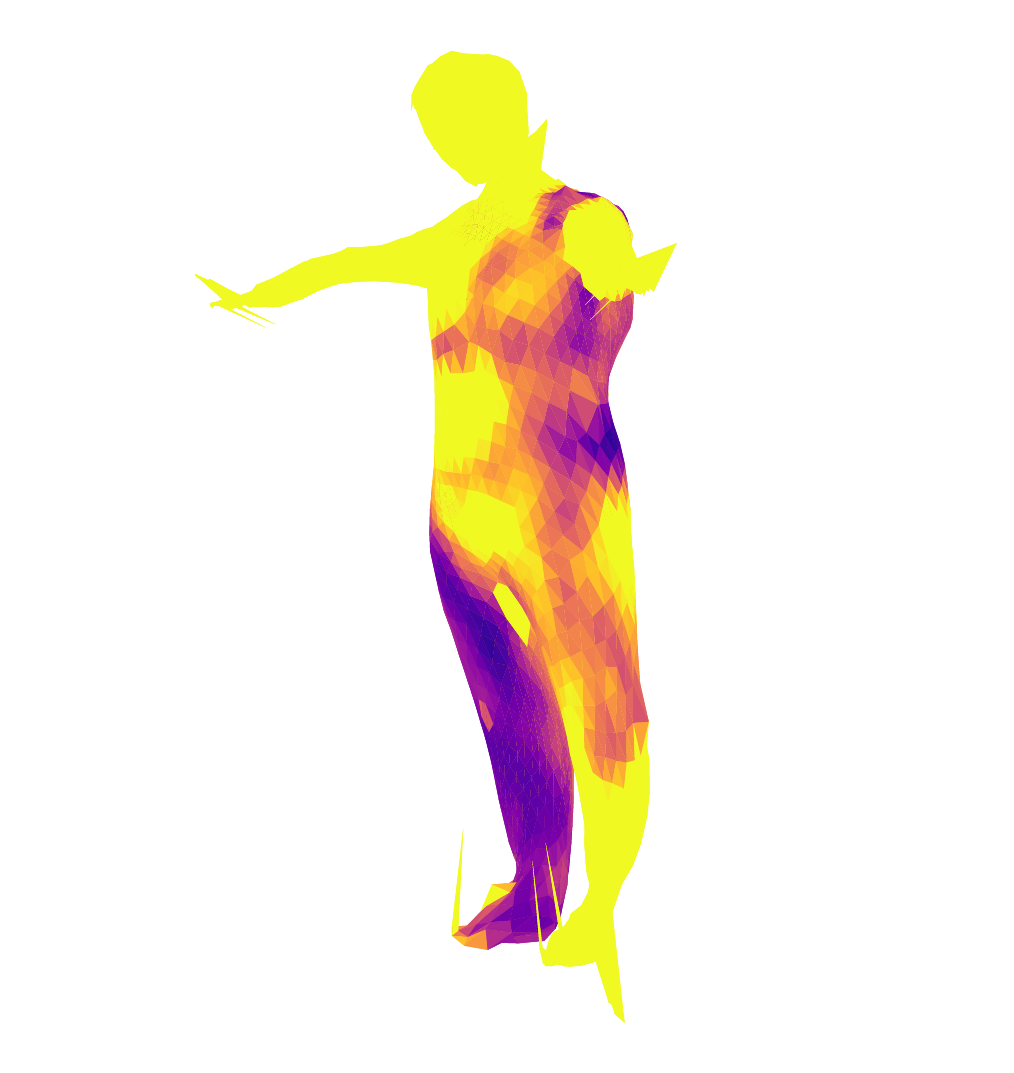}}} &
  \raisebox{-0.5\height}{{\includegraphics[width=0.8\linewidth, trim=13cm 12cm 28cm 15.2cm, clip]{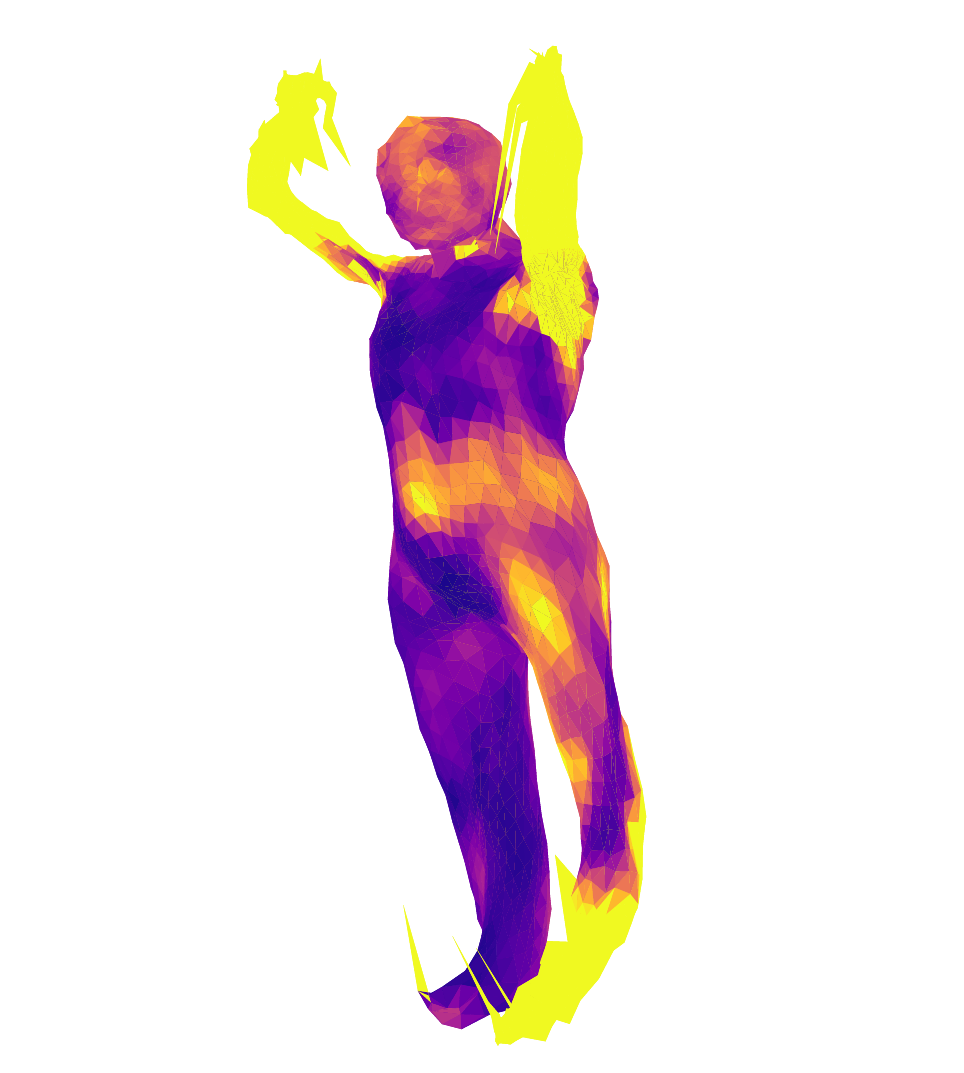}}} &
  \raisebox{-0.5\height}{{\includegraphics[width=0.8\linewidth, trim=13cm 12.5cm 28cm 14.5cm, clip]{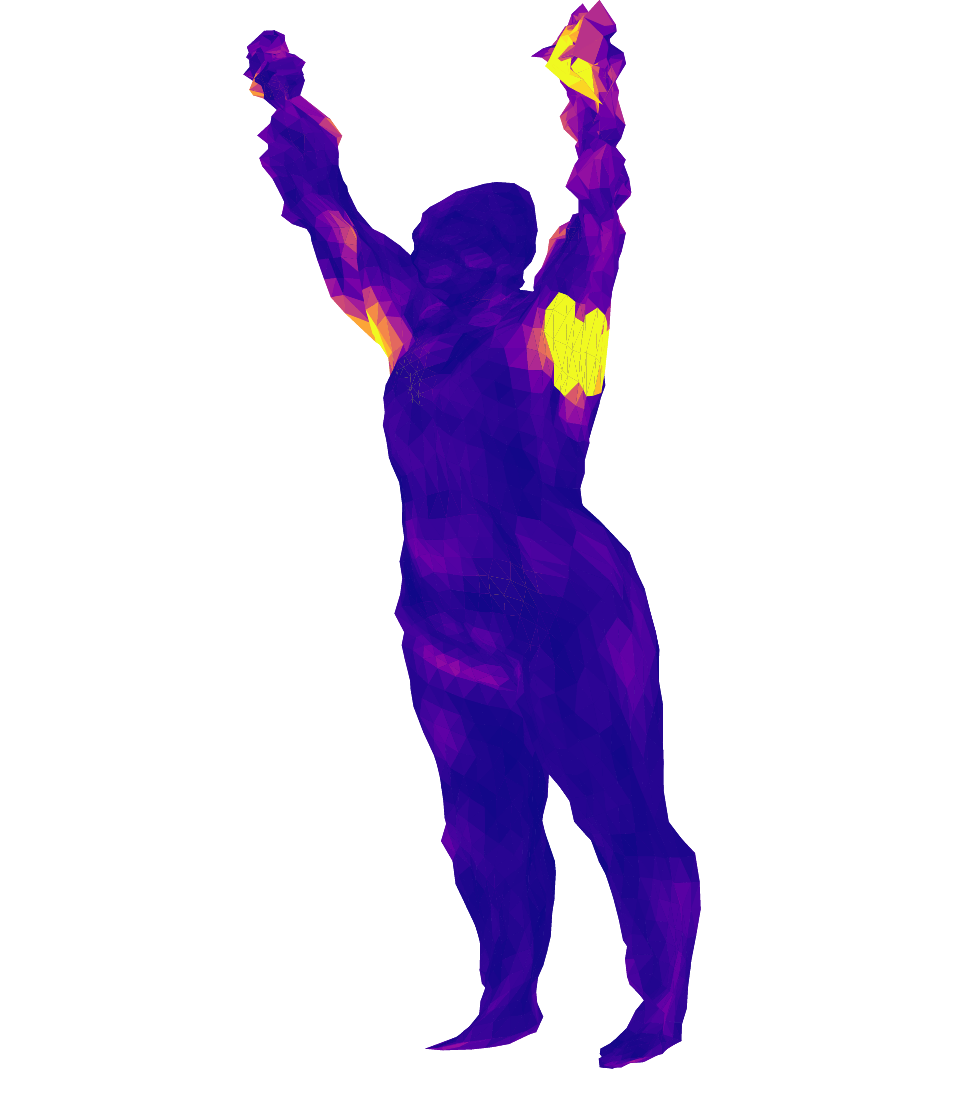}}}
  \\

  & & & \multicolumn{2}{c}{{\includegraphics[width=0.25\linewidth, trim=0cm .48cm 0cm 0.31cm, clip]{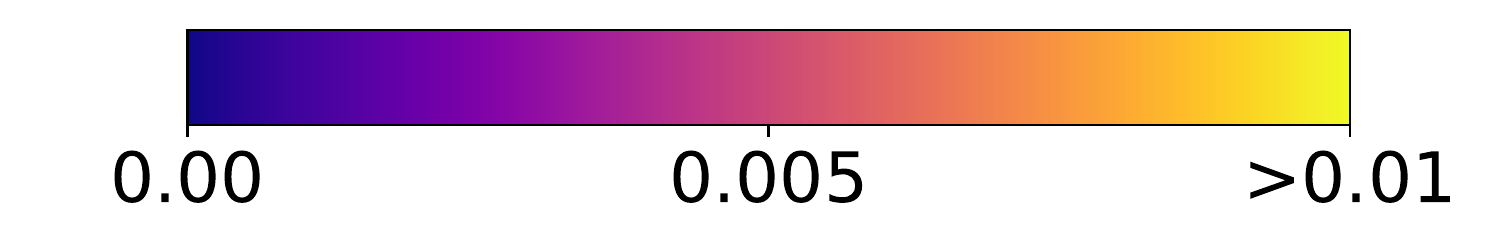}}}
 \end{tabular}
\end{minipage}
\end{center}
\caption{Reconstructed GALLOP and FAUST test samples by CoMA \cite{Ranjan2018}, Neural3DMM \cite{Bouritsas2019}, and our network. The mean squared error of the reconstructed faces is highlighted. More reconstruction examples are given in Figure \ref{reconstruction_app} in the supplementary material.}

\label{reconstruction}
\renewcommand{\arraystretch}{1}
\end{figure*}

\begin{table*}[t]
\centering
\begin{tabular}{|l|c|c|c|c|}
\hline
Mesh Class & \# Vertices & CoMA \cite{Ranjan2018} per mesh & Neural3DMM \cite{Bouritsas2019} per mesh & Ours \\ \hline \hline
FAUST & 3501 & & & \\
\hspace{2ex}known poses & & 0.0033 + 0.0058 & 0.00190 + 0.0037 & 0.00032 +0.0007 \\
\hspace{2ex}unknown poses & & 0.1963 + 0.3253 & 0.04233 + 0.0707 & 0.00054 + 00011 \\\hline  \hline
Horse & 3601& 0.00044 + 0.0015 & 0.00072 + 0.0023 & 0.00020 + 0.0003\\
Camel & 3467 & 0.00051 + 0.0018 & 0.00186 + 0.0116 & 0.00020 + 0.0003\\
\hline
Elephant & 3781 & 0.00088 + 0.0055 & 0.00321 + 0.0267 & \hspace{1ex} 0.00081 
+ 0.0160 $^\dag$\\ 
&  &  &  & \hspace{1ex}0.00155 + 0.0163$^*$\\
\hline 
\end{tabular}
\caption{
Mean squared errors of reconstructed unseen meshes and their standard deviations for two different training runs. We train one autoencoder for all three animals in the GALLOP dataset and one autoencoder for each experiment on the FAUST dataset. \\ 
$^*$:~the elephant has not been seen by the network during training. $^\dag$:~include elephant in training set.} \label{errors}
\end{table*}

\subsection*{Training Details}

We train the network 
(implemented in Pytorch \cite{Paszke2019}) with the adaptive learning rate optimization algorithm \cite{Kingma2015} using a learning rate of 0.001. For the GALLOP and the FAUST dataset we train for 500 epochs using a batch size of 100. For the TRUCK data we chose 250 epochs and a batch size of 50, since the variation inside the dataset is higher.
We minimize the mean squared error between original and reconstructed regional patches of the surface mesh without considering the padding.
To augment the data in the case of the GALLOP and the FAUST dataset we rotate the regional patches by 0$^{\circ}$, 120$^{\circ}$ and 240$^{\circ}$.

Figure \ref{anisotropickernel} shows trained 
hexagonal anisotropic kernels, which implies sensitivity to orientation.

\subsection*{Reconstructions of the Meshes}

\bgroup
\newcommand\cp{0.35}
\def\arraystretch{1}
\begin{table}[t]
\centering
\begin{tabular}{|@{\hspace{\cp em}}  l@{\hspace{\cp em}}  | @{\hspace{\cp em}}  c@{\hspace{\cp em}}  | @{\hspace{\cp em}}  c @{\hspace{\cp em}} | @{\hspace{\cp em}} c @{\hspace{\cp em}} |}
\hline
Mesh Class & CoMA \cite{Ranjan2018} & Neural3DMM \cite{Bouritsas2019} & Ours \\ \hline \hline
FAUST & 26795 & 276275 & 18184 \\
\hline \hline
Horse & 27339 & 280499 & \\
Camel & 26795 & 292659 & 18184\\
Elephant & 27339 & 296883 & \\ \hline 
\end{tabular}
\caption{Comparison of number of parameters. Our network requires at least 30\% fewer parameters.} 
\label{parameters}
\end{table}
\egroup

\begin{figure}
\renewcommand{\arraystretch}{0}
\begin{center}
\begin{minipage}{\linewidth}
  \centering
 \begin{tabular}{>{\centering\arraybackslash}p{0.45\linewidth}  >{\centering\arraybackslash}p{0.45\linewidth}}
 
  \textbf{Original Mesh} &  \textbf{Our Method's Reconstruction} \\

  \raisebox{-0.5\height}{{\includegraphics[width=\linewidth, trim=9.5cm 16.5cm 30cm 17cm, clip]{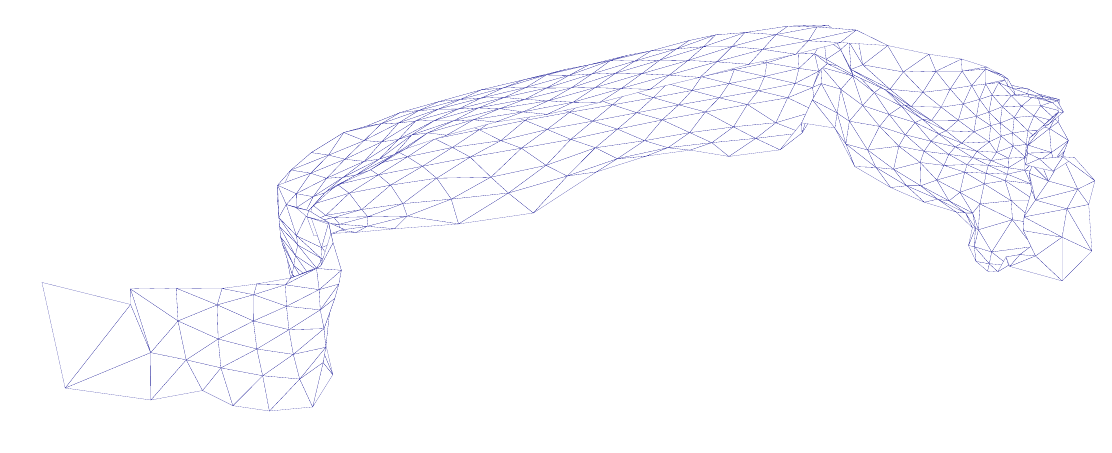}}} & 
  \raisebox{-0.5\height}{{\includegraphics[width=\linewidth, trim=9.5cm 16.5cm 30cm 17cm, clip]{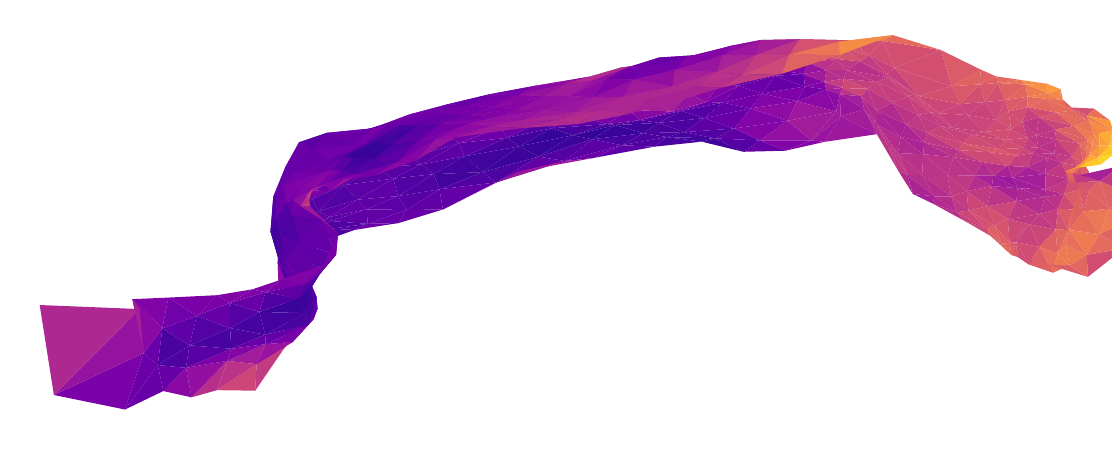}}} \\

  \raisebox{-0.5\height}{{\includegraphics[width=\linewidth, trim=9.5cm 16.8cm 30.5cm 17cm, clip]{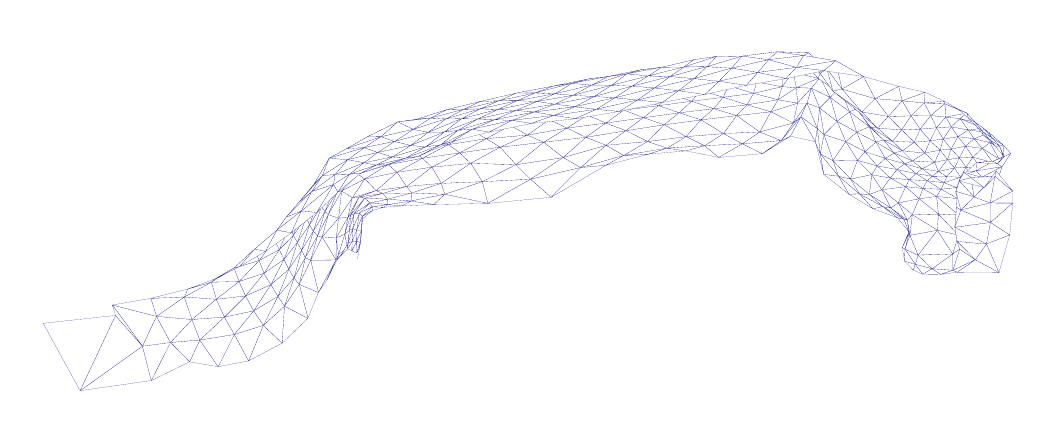}}} &
  \raisebox{-0.5\height}{{\includegraphics[width=\linewidth, trim=9.5cm 16.8cm 30.5cm 17cm, clip]{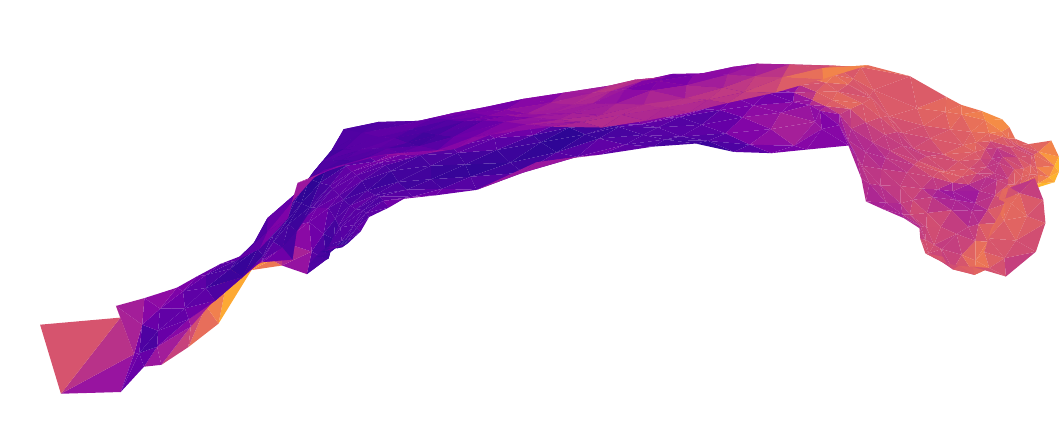}}}
  \\

 & \includegraphics[width=0.9\linewidth, trim=0cm .4cm 0cm 0.3cm, clip]{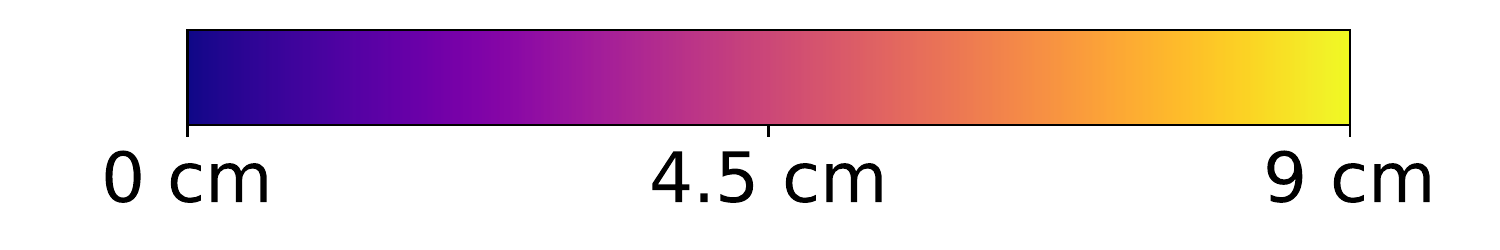}\\
 \end{tabular}
\end{minipage}
\end{center}
\caption{Reconstructed left front beam from the TRUCK (length of 150 cm) at time $t=24$ (test sample) from two different crash simulations. The average Euclidean distance (in cm) of the faces is highlighted. In the majority of faces the error is below 4.5 cm and only in highly deformed areas higher.}
\label{reconstruction_TRUCK}
\renewcommand{\arraystretch}{1}
\end{figure}

In Table \ref{errors} we compare our hexagonal mesh autoencoder to the CoMA \cite{Ranjan2018} and the Neural3DMM \cite{Bouritsas2019} network for the GALLOP and FAUST dataset in terms of the mean squared errors of reconstructed unseen shapes, whose 3D coordinates lie in range $[-1,1]$.
Albeit training the autoencoder for two different animal meshes of the GALLOP dataset, our network reduces the reconstruction error of unseen data by more than 50\%, if the animal is presented to the autoencoder during the training. 
We are also able to reconstruct a galloping sequence of an unseen elephant, although the reconstruction error is higher than with a baseline network trained only for this animal. If we include the elephant in the training set, the reconstruction errors are slightly lower compared to CoMA.
Especially the reconstruction of the legs is superior with our method in comparison to CoMA and Neural3DMM, as Figure \ref{reconstruction} and Figure \ref{reconstruction_app} in the supplementary material illustrate.
Note that the mesh reconstruction is smooth at the patches' boundaries. This indicates that the padding enforces the incorporation of information from the neighboring patches.

Our network reconstructs known and unknown poses of FAUST with a more than 80\% lower error, see Table \ref{errors} and Figure \ref{reconstruction_app} in the supplementary material. Limbs are reconstructed inaccurately by the CoMA and Neural3DMM architectures. Especially if the pose is unknown and not similar to training poses, their reconstruction fails.

In all cases, our architecture requires fewer parameters than the CoMA and Neural3DMM 
networks (Table \ref{parameters}).

\bgroup
\newcommand\cp{0.35}
\def\arraystretch{1}
\begin{table}[t]
\centering
\begin{tabular}{|@{\hspace{\cp em}} l @{\hspace{\cp em}}| @{\hspace{\cp em}}c @{\hspace{\cp em}} | @{\hspace{\cp em}} c @{\hspace{\cp em}} | @{\hspace{\cp em}} c @{\hspace{\cp em}}|}
\hline
Dataset & Train MSE & Test MSE & Eucl. E. \\ \hline \hline
TRUCK & 0.0026 + 0.004 & 0.0027 + 0.004 & 5.49 $|$ 3.76 \\\hline
YARIS & -- & 0.0147 + 0.023 & 2.23 $|$ 1.92 \\
\hline
\end{tabular}
\caption{Mean squared errors of reconstructed unseen meshes and their standard deviations for two different training runs on the TRUCK and YARIS dataset. Additionally, the average Euclidean vertex wise error and its median (in cm) are given.} \label{errors_car}
\end{table}
\egroup

Since the TRUCK and YARIS datasets contain 16 different meshes, we would have to train the baseline architectures 16 times. To validate the reconstruction results, we here calculate the error in cm  and have a closer look at the embeddings.
In Table \ref{errors_car} we present the mean squared errors and average Euclidean vertex wise error and median for the TRUCK dataset, for which the autoencoder has been trained, and for the YARIS dataset that has not been presented to the network during training. The TRUCK components measure between 135 and 370 cm in length, the YARIS components between 21 and 91 cm.
If we put the error in relation to the part length, the average Euclidean error on the unseen YARIS dataset is less than 3 times higher in comparison to the testing error on the TRUCK, for which selected reconstruction results are presented in Figure \ref{reconstruction_TRUCK}.

\subsection*{Low-dimensional Embedding}

\begin{figure}
\begin{center}
\includegraphics[width=0.82\linewidth]{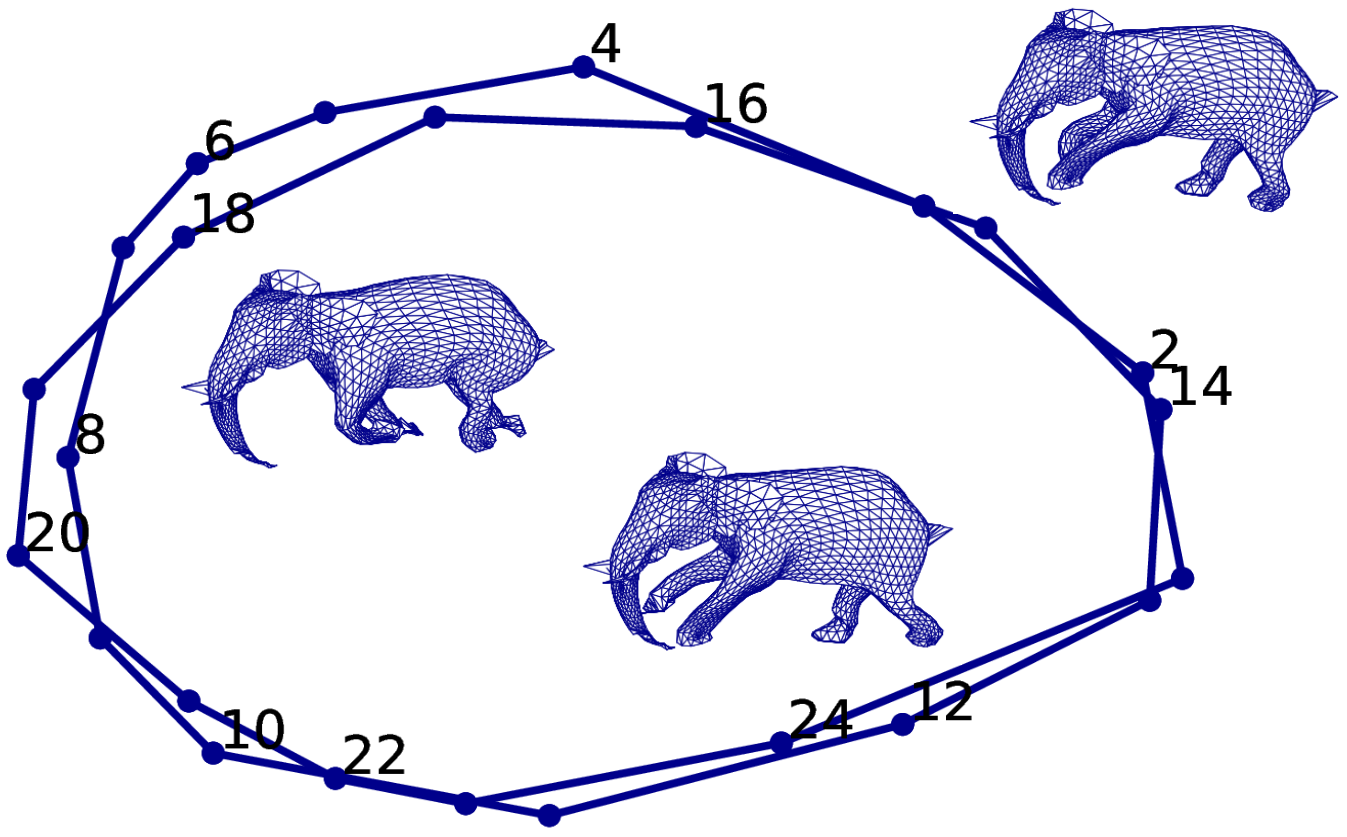}
\end{center}
\caption{Embedded cyclic sequence of the galloping elephant, whose mesh has not been seen during training.}
\label{embedding}
\end{figure}

For the three different animals from the GALLOP dataset we can visualize the low-dimensional embeddings of the galloping sequences of the three animals.
We concatenate the patch wise embeddings for each timestep and project the resulting vectors to the two-dimensional space using Principal Component Analysis (PCA) \cite{Pearson01}.
The time dependent embeddings for all three animals, whether included in the training or only in the testing set, exhibit a periodic galloping sequence in the two-dimensional space, as seen in Figure \ref{embedding} for the elephant's embedding. 
The embeddings of the cyclic sequences are similar to the ones from \cite{Yuan2020}, who trained one autoencoder for each animal and showed that the direct application of PCA or t-SNE cannot reveal the intrinsic information  of the data.

\begin{figure}
\begin{center}
\includegraphics[width=0.95\linewidth, trim=0cm 0cm 0cm 0cm, clip]{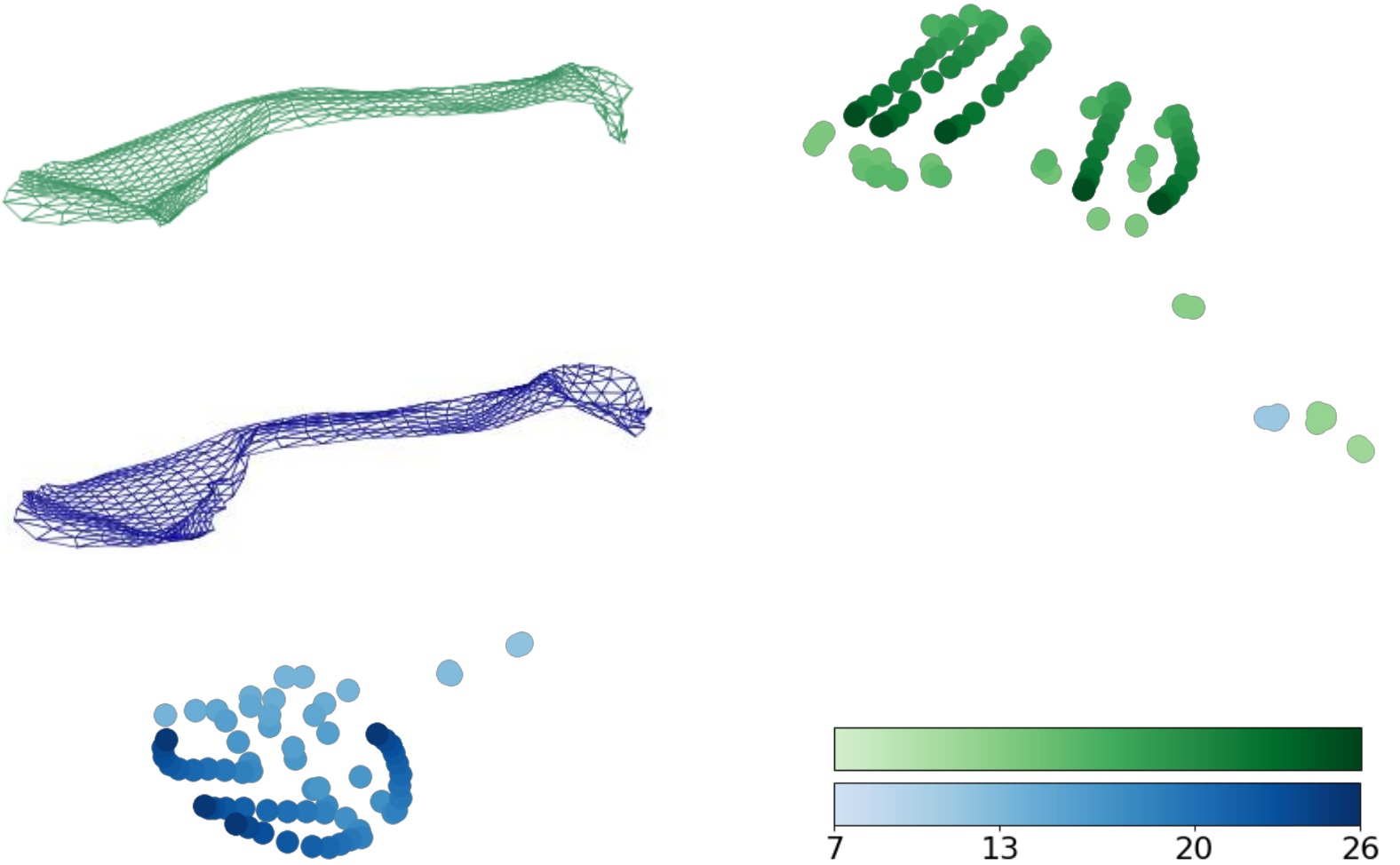}
\end{center}
\caption{Embedded low-dimensional representation of the front beam of the YARIS. For the timesteps $t= 7,\dots,26$ we observe two patterns in the deformation behavior. The mesh has not been presented to the autoencoder during training.}
\label{emb_yaris}
\end{figure}

For each car component we create a 2D-visualization of the low-dimensional representation using t-SNE \cite{vanderMaaten2008} to detect patterns in the deformation, that separates the simulations into clusters. This speeds up the analysis of car crash simulations, since relations between model parameters and the deformation behavior are discovered faster \cite{Bohn2013,Hahner2020}.
For the TRUCK we observe that the selected components deform in two different branches for the 32 simulations. This behavior manifests after approximately half of the time similar to \cite{Bohn2013,Hahner2020}.
Figure \ref{emb_truck} in the supplementary material visualizes this for the left front beam.
For the YARIS, which has never been seen by the network during training, we visualize the low-dimensional representation in 2D using t-SNE \cite{vanderMaaten2008}. 
We detect a deformation pattern in the front beam that splits up the simulation set into two clusters, see Figure \ref{emb_yaris}.

\section{Conclusion}

We have introduced a novel approach for the analysis of deforming 3D surface meshes with a mesh autoencoder for semi-regular meshes. To the best of our knowledge, the remeshing of the triangular meshes into semi-regular ones allows for the first time an analysis of shapes of different size and geometry. The regular local structure makes a reutilization of the learned convolutional filters and an efficient mesh-independent pooling operation possible.

We evaluate our network successfully on three datasets from different domains and reconstruct the meshes in significantly better quality than the baseline mesh autoencoders. Additionally, we apply our trained autoencoder to unseen meshes of different shape and connectivity and successfully detect the underlying dynamics of unseen time sequences. 
In future work we plan to apply the architecture to other tasks such as shape matching and segmentation.

{\small
\bibliographystyle{ieee_fullname}
\bibliography{210817_mendeley2.bib}
}

\cleardoublepage

\newpage

\section*{Supplementary Material}

\subsection*{Remeshing Algorithm}

At first, a coarse approximation of the input mesh is built. To coarsen the surface meshes we employ the  Garland-Heckbert-algorithm for surface simplification using quadric error metrics \cite{Garland1997}.
It simplifies the mesh by collapsing edges until the target number of faces is reached, always contracting the pair of edges with the lowest cost. The cost measures the shape changes. As in \cite{Yuan2020}, we regularize the edge lengths to have regularly distributed vertices. Additionally, we prevent the algorithm from contracting edges, that lead to non-manifold edges in the mesh.

The coarse approximation of the input mesh is subdivided to the desired level of subdivision. 

Now, the resulting semi-regular mesh $\mathcal{M}_{SR}$ has to be fit to the original irregular mesh $\mathcal{M}_{IR}$ in order to describe the surface well.

We chose stochastic gradient descent to optimize a loss function that describes how well the semi-regular mesh fits to the irregular mesh. The loss function is optimized with respect to a deformation vector, that contains a 3D offset for each vertex of the semi-regular mesh $\mathcal{M}_{SR}$.
The loss function employs the average chamfer distance between sampled points $S_{IR}$ from the surfaces described by the original mesh $\mathcal{M}_{IR}$ and sampled points $S_{SR}$  from the iteratively deformed semi-regular mesh $M_{SR}$ respectively. 
We use the following definition of the chamfer distance \cite{Achlioptas2018} which measures the average squared distance between each point in set $S_{IR}$ to its nearest neighbor in the other set $S_{SR}$.

\begin{align*}
    d_{avgCD}(S_{IR}, S_{SR}) = & \frac{1}{|S_{IR}|} \sum_{x\in S_{IR}} \min_{y\in S_{SR}} \lVert x-y \rVert_2^2 + \\
    & \frac{1}{|S_{SR}|} \sum_{y\in S_{SR}} \min_{x\in S_{IR}} \lVert x-y \rVert_2^2
\end{align*}

Additionally, we regularize the lengths of the edges of $\mathcal{M}_{SR}$, smooth the Laplacian of $\mathcal{M}_{SR}$ and enforce consistency across the normals of neighboring faces of $\mathcal{M}_{SR}$. The regularization terms are weighted. 
To fit the semi-regular mesh to the original irregular mesh we utilize an implementation in Pytorch3D \cite{Ravi2020} that is based on  \cite{Ravi2020}\footnote{\url{https://pytorch3d.org/tutorials/deform_source_mesh_to_target_mesh}}.

Note that if for a deforming shape the mesh topology stays constant over time, one can just remesh one undeformed template mesh. The semi-regular remeshing result is parameterized and transferred to the meshes at the different timesteps, which describe the same shape.
For that, after projecting the vertices of the semi-regular mesh to the closest face of the irregular template mesh, we can calculate the barycentric coordinates and obtain a parametrization. This parametrization of the remeshing result can be applied to the other deformed meshes and the complete sequence of the deforming shape is discretized by semi-regular meshes. Note that this is for simplification of the overall workflow, and for ease of visualization of the galloping sequences. In principle, a parametrization can be calculated for every timestep between the irregular mesh and the semi-regular mesh.

\subsection*{Tables and Figures}

As an addition to the architecture's description in section \ref{sec:autoencoder} and visualization in Figure \ref{pooling} we give a detailed distribution of parameters over the hexagonal convolutional, fully connected, and pooling layers in Table \ref{network}.

\begin{table}[!h]
\centering
\begin{tabular}{|l|c|c|c|}
\hline
Layer & Output Shape & KS & Param.  \\ \hline \hline
Input & $({\scriptstyle \bullet},3,111)$ & & 0 \\
HexConv & $({\scriptstyle \bullet}, 16, 111)$ & 2 & 912  \\
Pooling & $({\scriptstyle \bullet}, 16, 33)$ &  & 0 \\
HexConv  & $({\scriptstyle \bullet}, 32, 33)$ & 1 & 3584 \\
Pooling & $({\scriptstyle \bullet}, 32, 6)$ &  & 0  \\
Fully Connected & $({\scriptstyle \bullet}, 8)$ & & 2312  \\ \hline 
 \multicolumn{4}{|c|}{Hidden Representation for each patch of size 8} \\ \hline
Fully Connected & $({\scriptstyle \bullet}, 32, 6)$ & & 2592  \\ 
Unpooling & $({\scriptstyle \bullet},32, 33)$ & & 0  \\
HexConv & $({\scriptstyle \bullet}, 16, 33)$ & 1 & 3584 \\
Unpooling & $({\scriptstyle \bullet},16, 111)$ & & 0  \\
HexConv & $({\scriptstyle \bullet}, 16, 111)$ & 2 & 4864 \\
HexConv & $({\scriptstyle \bullet}, 3, 111)$ & 1 & 336  \\
\hline
\end{tabular}
\caption{Structure of the autoencoder. The bullets ${\scriptstyle \bullet}$ reference the corresponding batch size. The data's last dimension is the number of vertices considered for each padded patch. For hexagonal convolutional layers the kernel size (KS) is given. } \label{network}
\end{table}

\begin{figure*}
\renewcommand{\arraystretch}{0}
\setlength{\tabcolsep}{0em}
\begin{center}
\begin{minipage}{\linewidth}
  \centering
 \begin{tabular}{p{0.07\linewidth} 
 >{\centering\arraybackslash}p{0.22\linewidth}  
 >{\centering\arraybackslash}p{0.237\linewidth}
 >{\centering\arraybackslash}p{0.237\linewidth}
 >{\centering\arraybackslash}p{0.237\linewidth}}
 
  &\textbf{Original Mesh} & \textbf{CoMA \cite{Ranjan2018} } & \textbf{Neural3DMM \cite{Bouritsas2019} }& \textbf{Our Reconstruction} \\

  Horse $t=43$ & \raisebox{-0.5\height}{{\includegraphics[width=\linewidth, trim=12cm 16.5cm 28cm 15.5cm, clip]{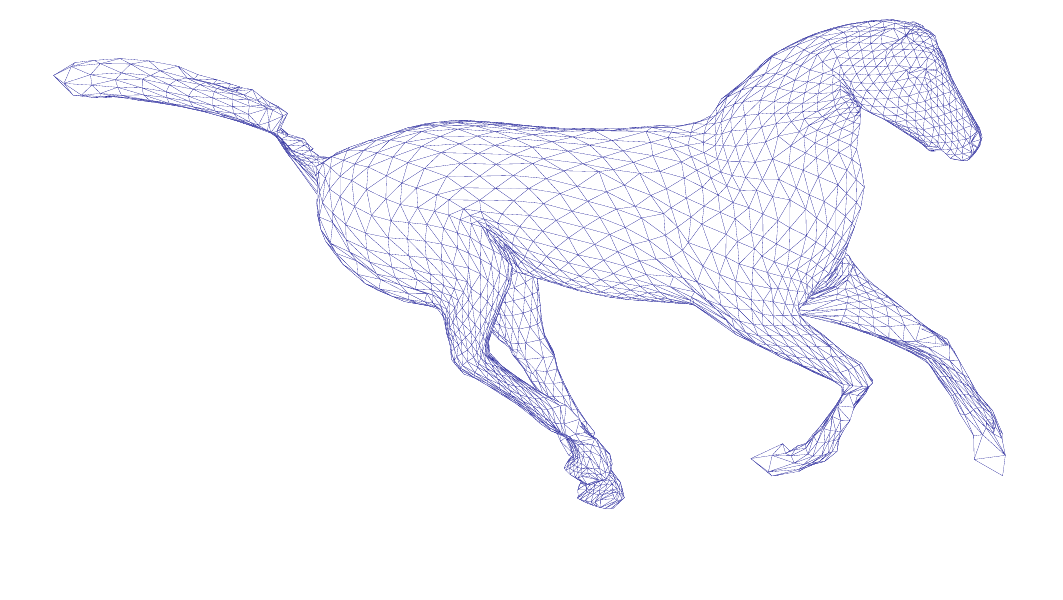}}} &
  \raisebox{-0.5\height}{{\includegraphics[width=\linewidth, trim=12cm 17.5cm 28cm 15.5cm, clip]{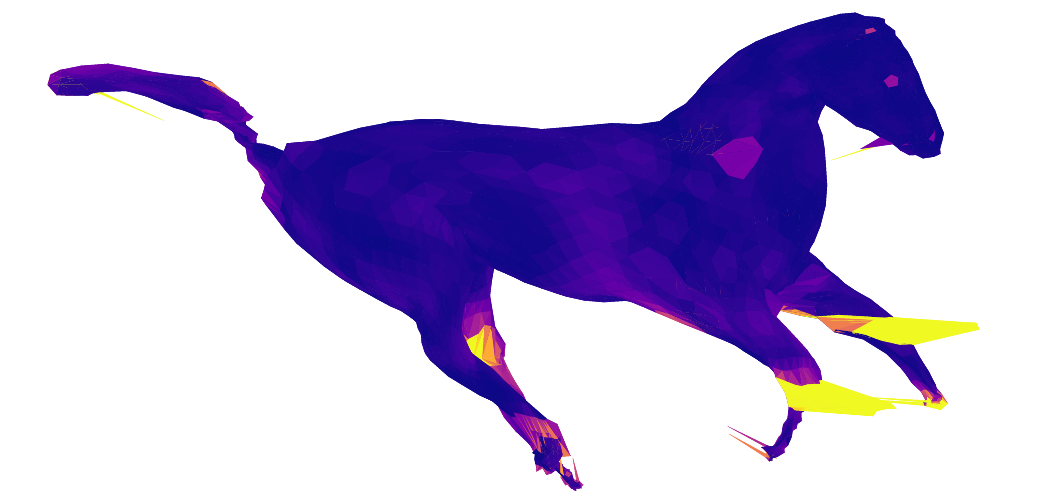}}} &
  \raisebox{-0.5\height}{{\includegraphics[width=\linewidth, trim=12cm 17.5cm 28cm 15.5cm, clip]{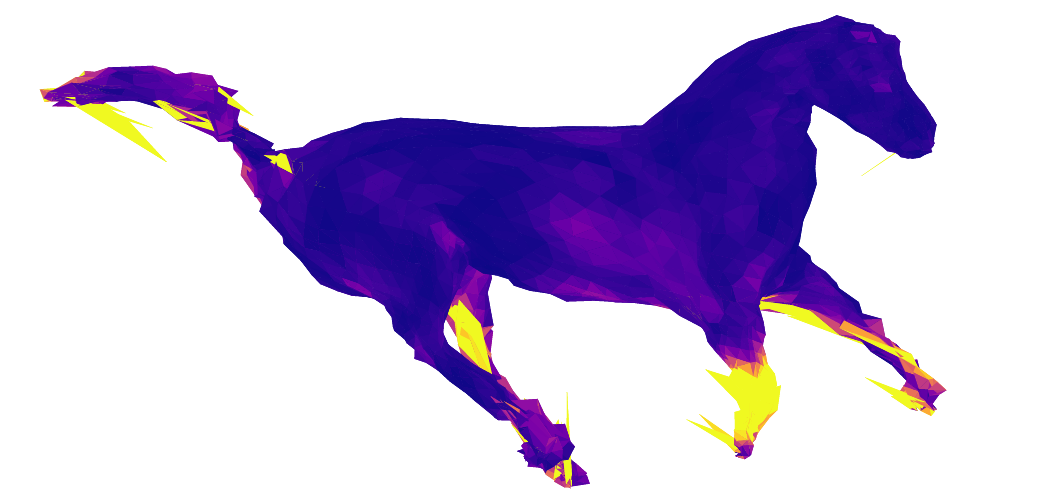}}} &
  \raisebox{-0.5\height}{{\includegraphics[width=\linewidth, trim=12cm 17.5cm 28cm 15.5cm, clip]{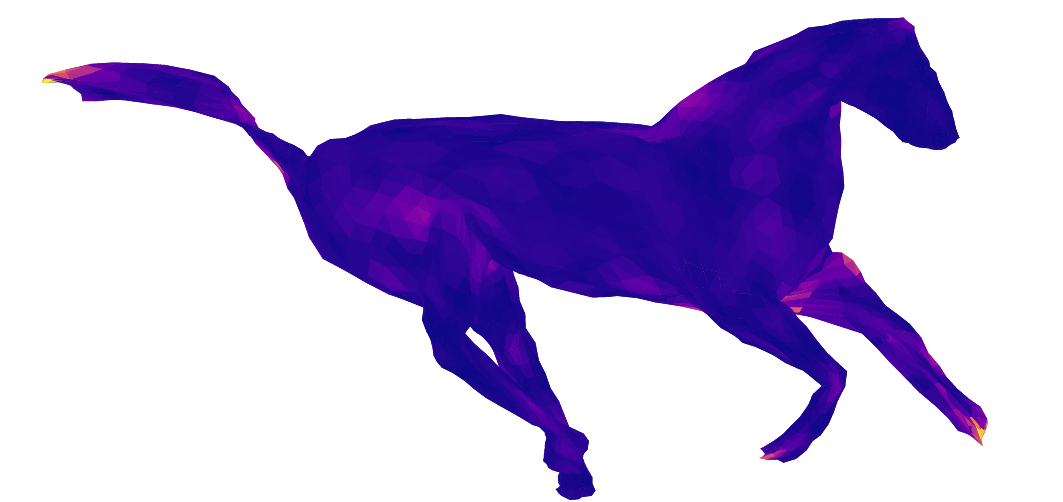}}} \\

  Camel $t=43$ & \scalebox{-1}[1]{\raisebox{-0.5\height}{{\includegraphics[width=\linewidth, trim=14cm 15cm 27cm 17cm, clip]{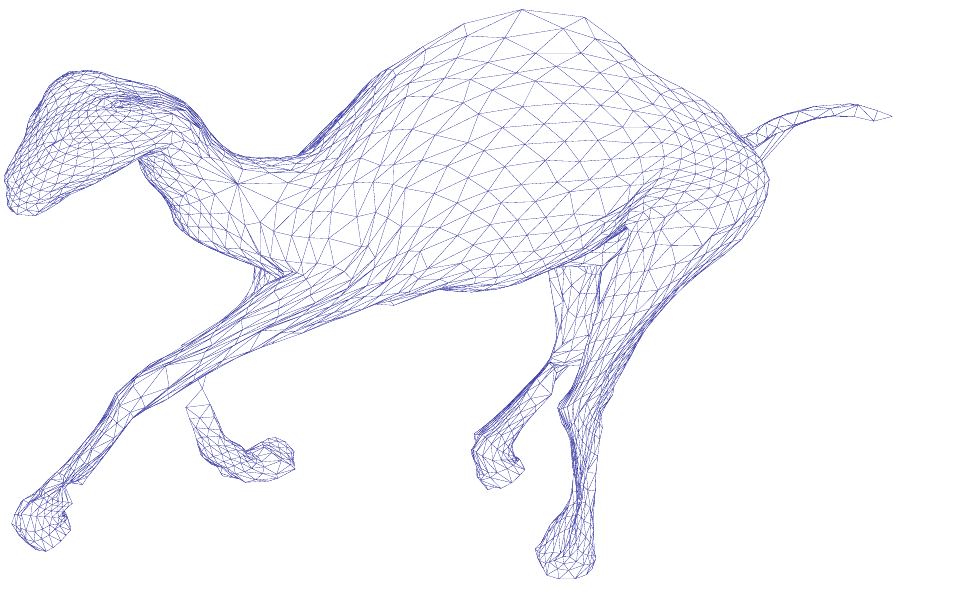}}}} &
  \scalebox{-1}[1]{\raisebox{-0.5\height}{{\includegraphics[width=\linewidth, trim=13cm 15cm 27cm 17cm, clip]{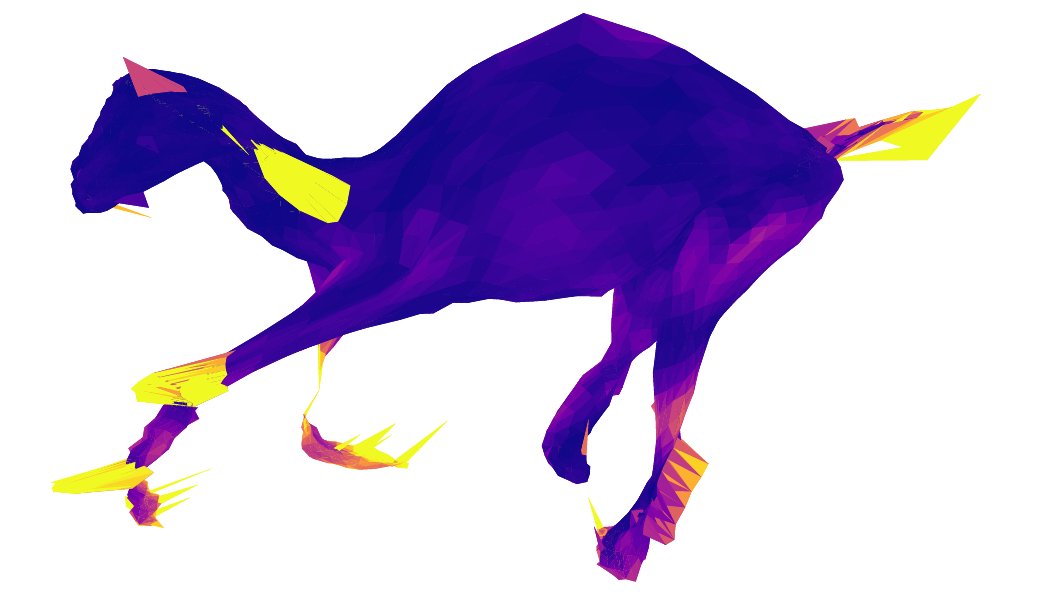}}}} &
  \scalebox{-1}[1]{\raisebox{-0.5\height}{{\includegraphics[width=\linewidth, trim=13.5cm 15cm 27cm 17cm, clip]{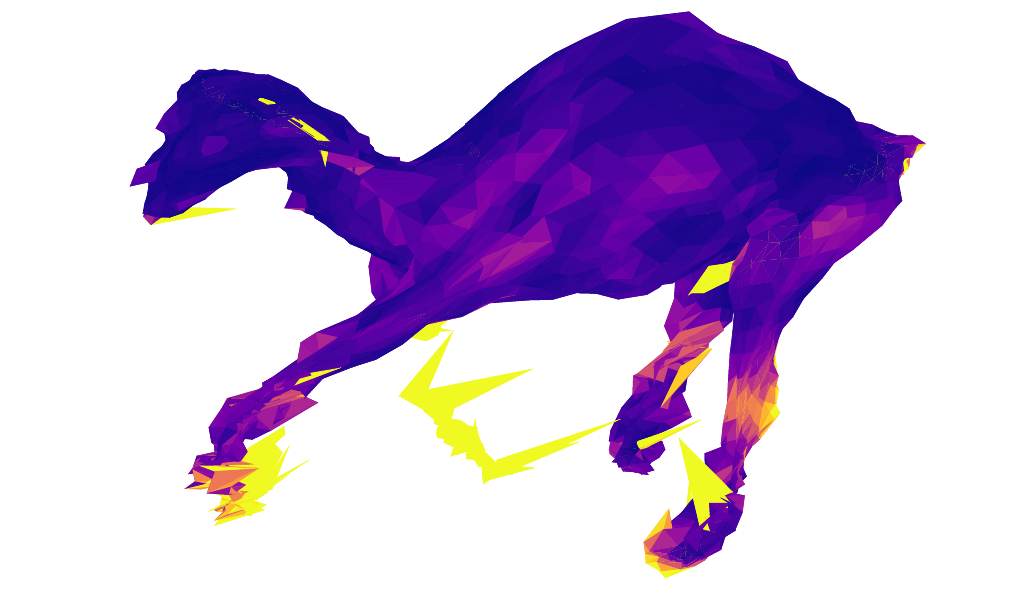}}}} &
  \scalebox{-1}[1]{\raisebox{-0.5\height}{{\includegraphics[width=\linewidth, trim=13.5cm 15cm 27cm 17cm, clip]{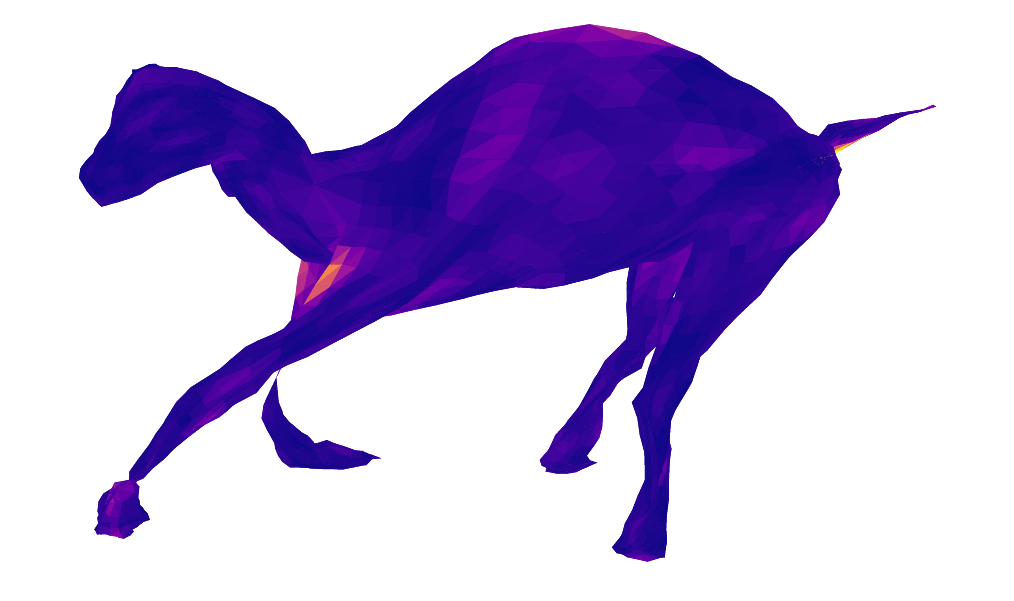}}}} \\

  FAUST known pose & 
  \raisebox{-0.5\height}{{\includegraphics[width=\linewidth, trim=11cm 11.5cm 28cm 15cm, clip]{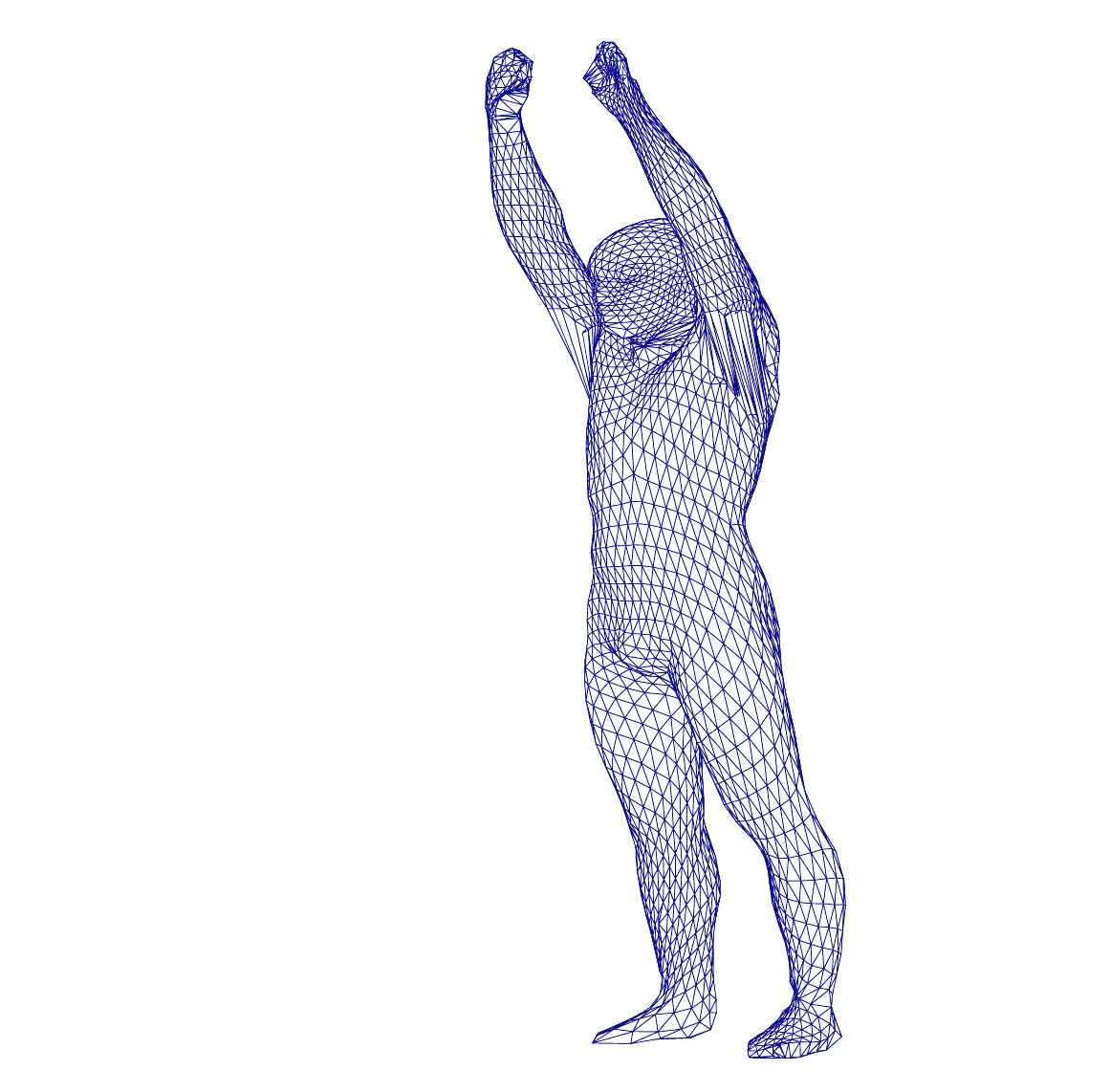}}} &
  \raisebox{-0.5\height}{{\includegraphics[width=\linewidth, trim=11cm 12cm 28cm 15.2cm, clip]{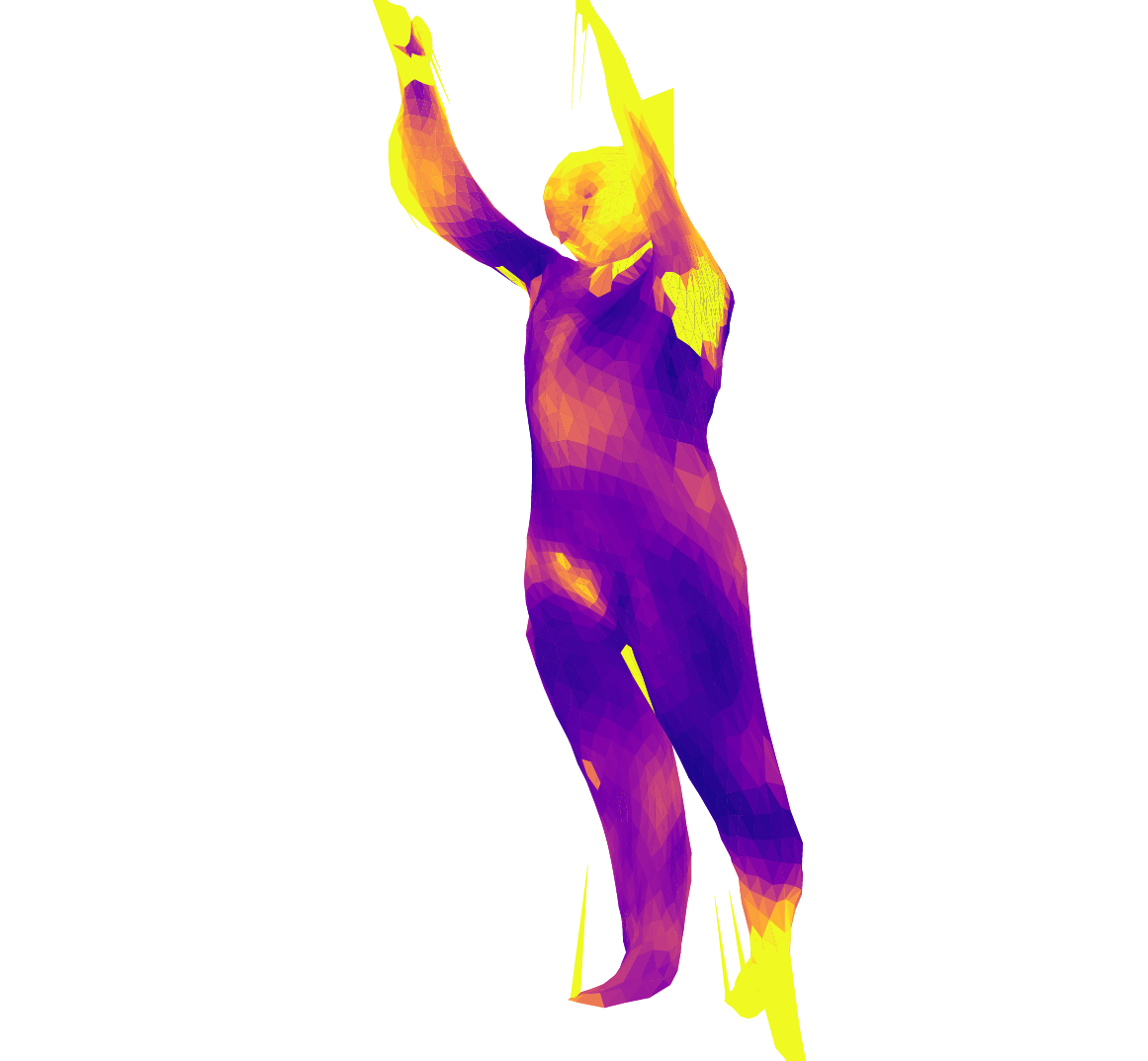}}} &
  \raisebox{-0.5\height}{{\includegraphics[width=\linewidth, trim=11cm 12cm 28cm 15.2cm, clip]{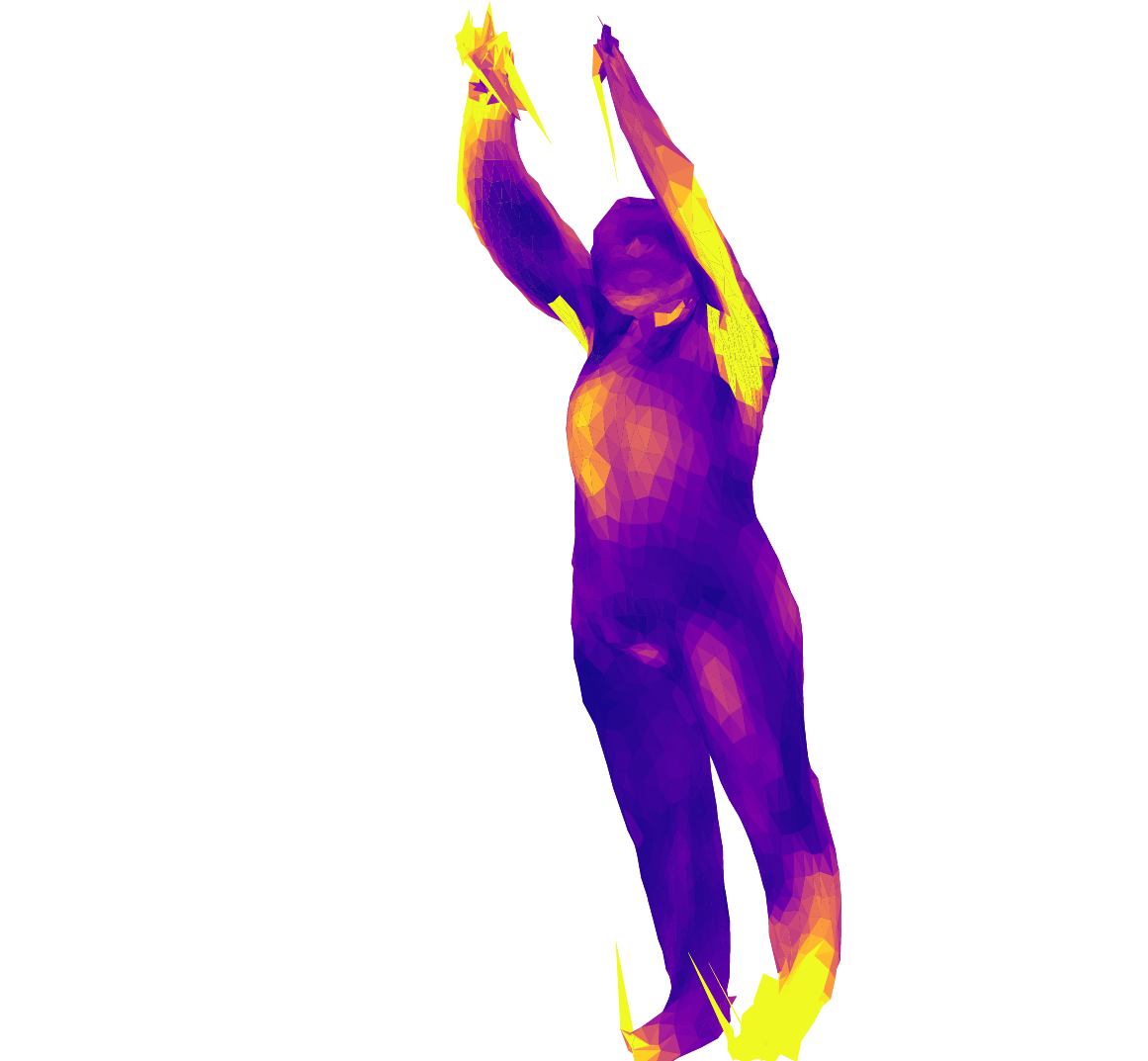}}} &
  \raisebox{-0.5\height}{{\includegraphics[width=\linewidth, trim=11cm 11.8cm 28cm 15.2cm, clip]{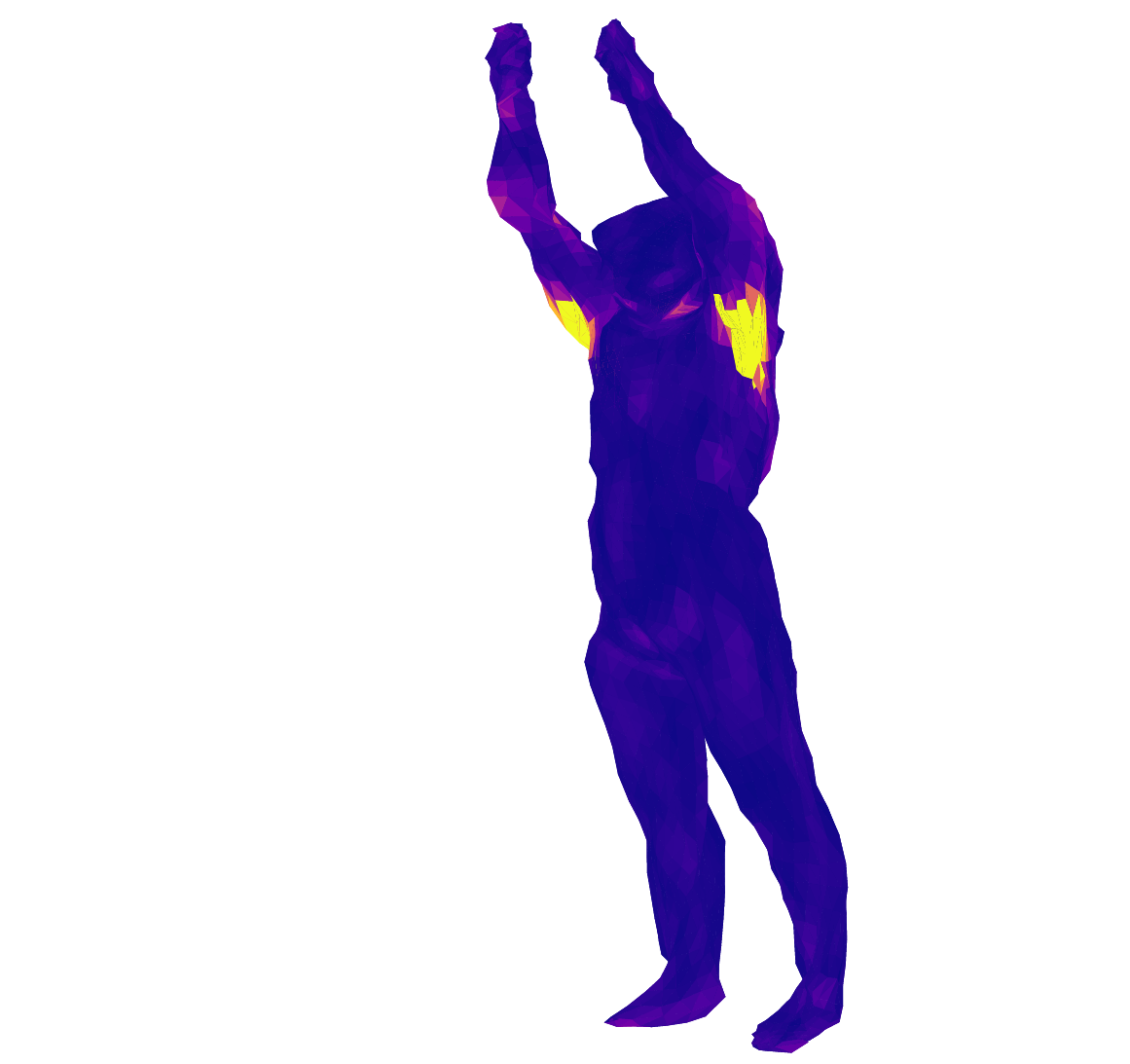}}}
  \\
  
  FAUST \mbox{unknown} pose & 
  \raisebox{-0.5\height}{{\includegraphics[width=0.95\linewidth, trim=11cm 12.5cm 28.5cm 14cm, clip]{output_plots/time_19_true_part_faust1_unknown_pose.pdf}}} &  
  \raisebox{-0.5\height}{{\includegraphics[width=\linewidth, trim=10.5cm 12.2cm 28cm 15cm, clip]{output_plots/COMA_time_19_predicted_part_faust4_unknownpose.pdf}}} &
  \raisebox{-0.5\height}{{\includegraphics[width=\linewidth, trim=10.5cm 12cm 28cm 15.2cm, clip]{output_plots/Neural3DMM_faust_dim01_time_19_predicted_pose_9_body_faust1.pdf}}} &
  \raisebox{-0.5\height}{{\includegraphics[width=\linewidth, trim=10.5cm 12.5cm 28cm 14.5cm, clip]{output_plots/time_19_predicted_part_faust1_unknown_pose_.pdf}}}
  \\
   
  FAUST \mbox{unknown} pose & 
  \raisebox{-0.5\height}{{\includegraphics[width=0.95\linewidth, trim=11.5cm 12.5cm 28.5cm 16cm, clip]{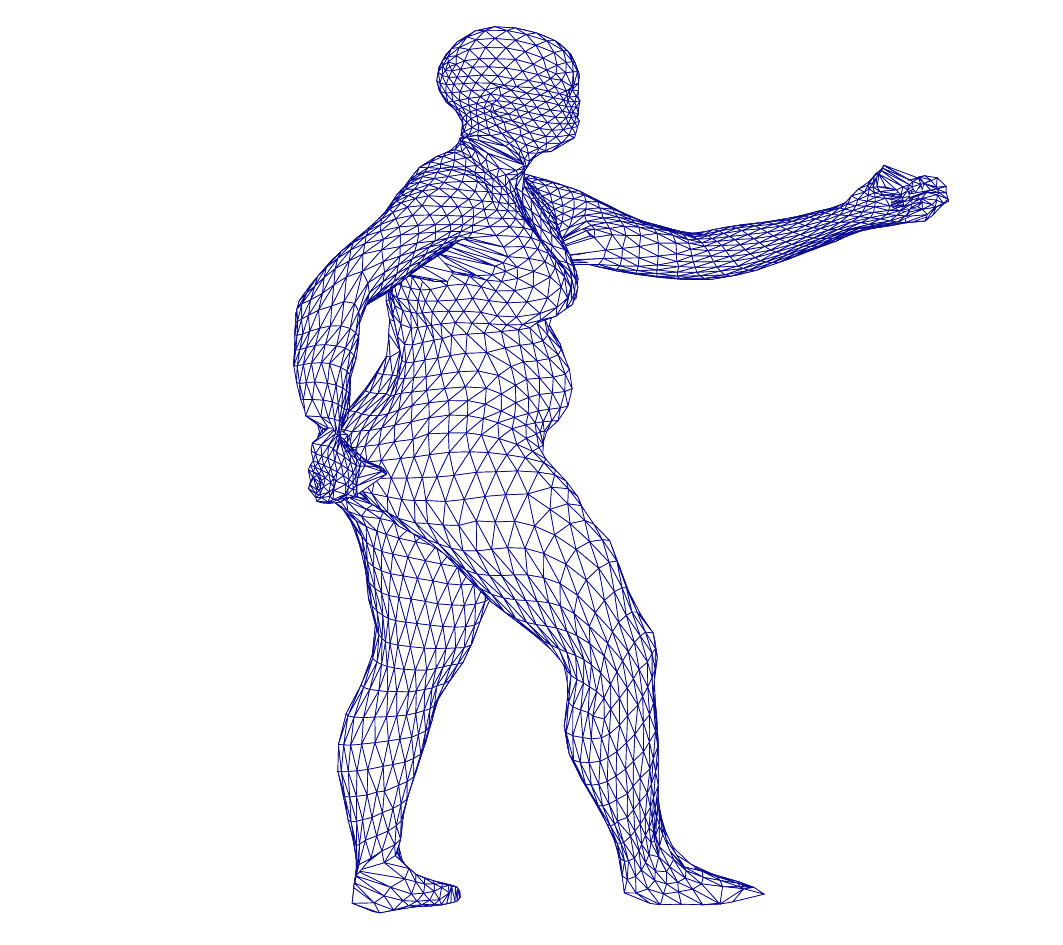}}} &
  \raisebox{-0.5\height}{{\includegraphics[width=\linewidth, trim=14cm 11.2cm 26cm 17.5cm, clip]{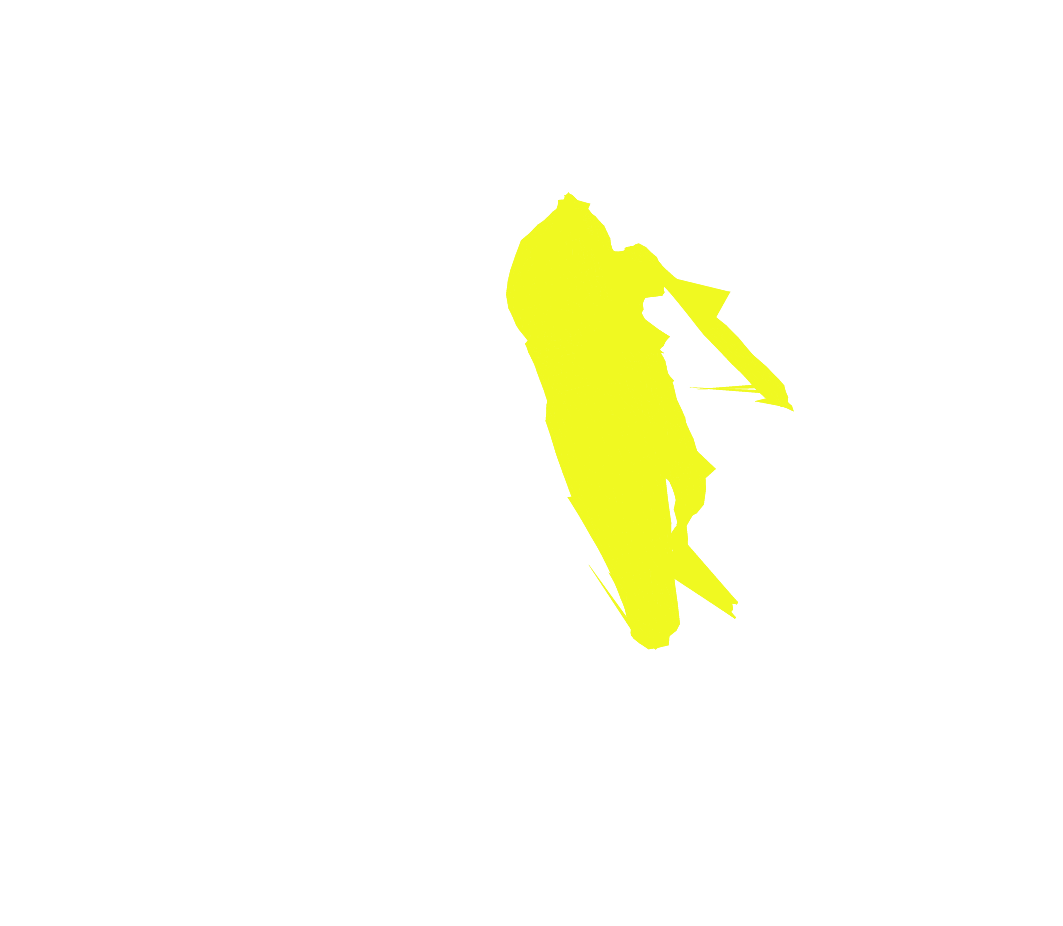}}} &
  \raisebox{-0.5\height}{{\includegraphics[width=\linewidth, trim=11cm 11.2cm 28cm 17.5cm, clip]{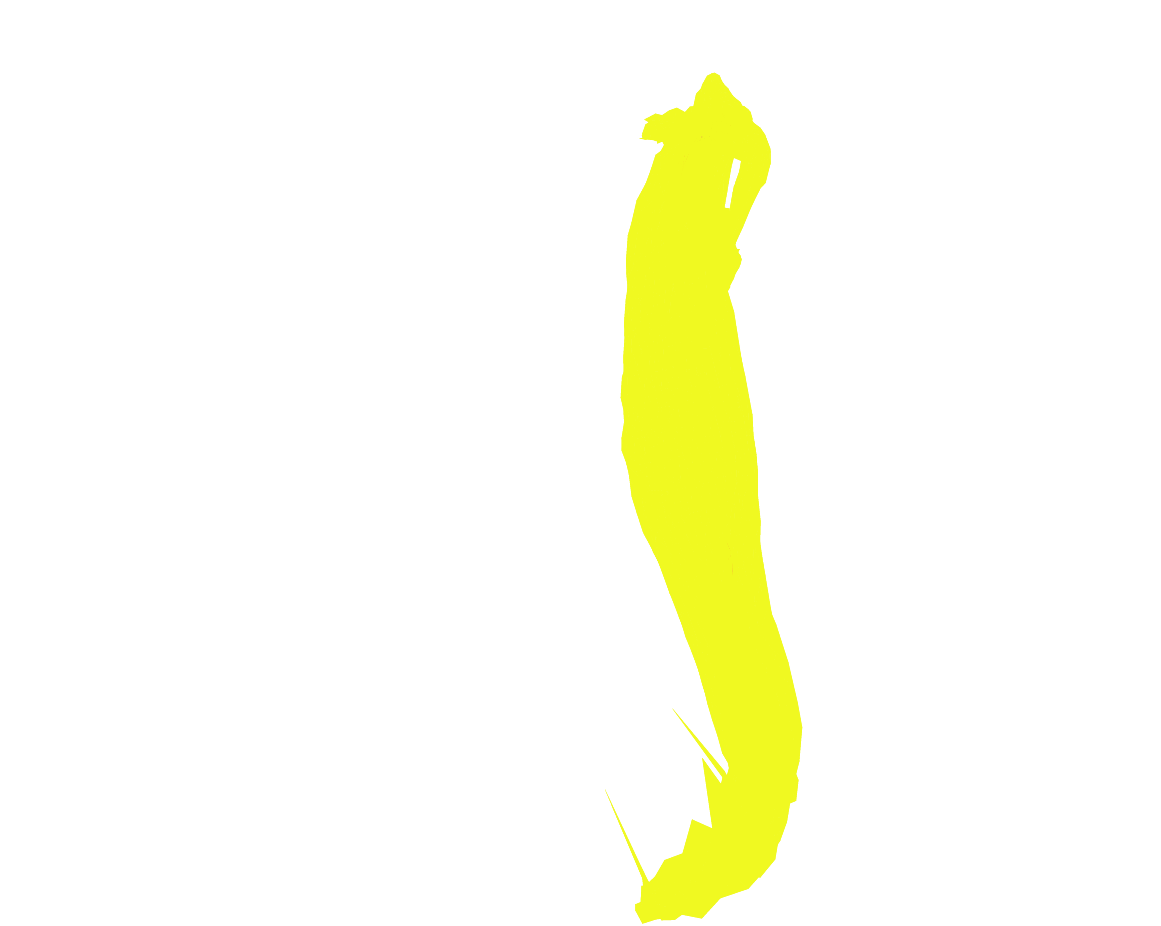}}} &
  \raisebox{-0.5\height}{{\includegraphics[width=\linewidth, trim=11cm 12.5cm 28cm 16cm, clip]{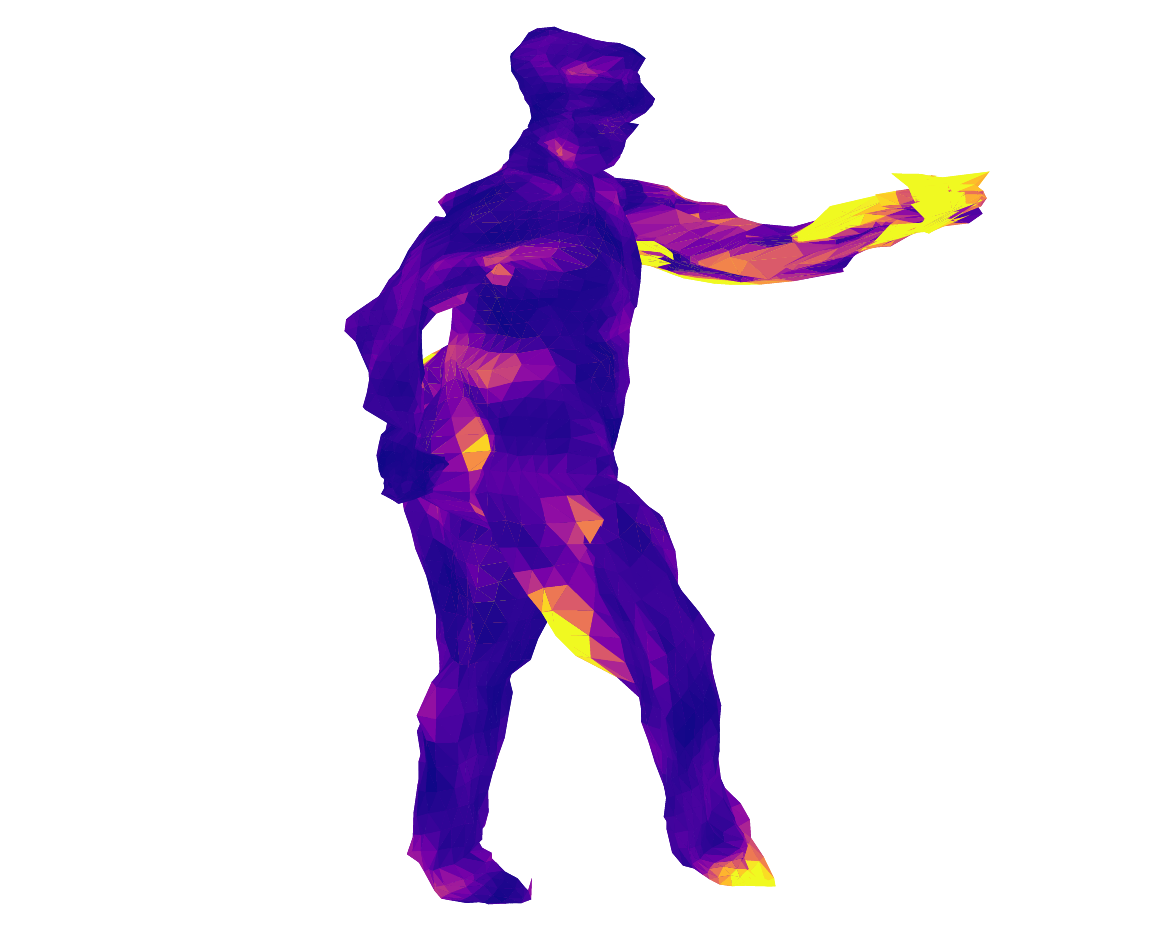}}}
  \\

  & & & \multicolumn{2}{c}{\includegraphics[width=0.25\linewidth, trim=0cm .4cm 0cm 0.3cm, clip]{output_plots/colorbar_part_horse.pdf}}
 \end{tabular}
\end{minipage}
\end{center}
\caption{Additional reconstructed GALLOP and FAUST test samples by CoMA \cite{Ranjan2018}, Neural3DMM \cite{Bouritsas2019}, and our network. The mean squared error of the reconstructed faces is highlighted.}
\label{reconstruction_app}
\end{figure*}

Figure \ref{reconstruction_app} shows more reconstruction results of our architecture and the baseline CoMA \cite{Ranjan2018} and Neural3DMM \cite{Bouritsas2019} autoencoder on test samples from the GALLOP and FAUST dataset.

\begin{figure}
\begin{center}
\includegraphics[width=0.99\linewidth, trim=0cm 0cm 0cm 0cm, clip]{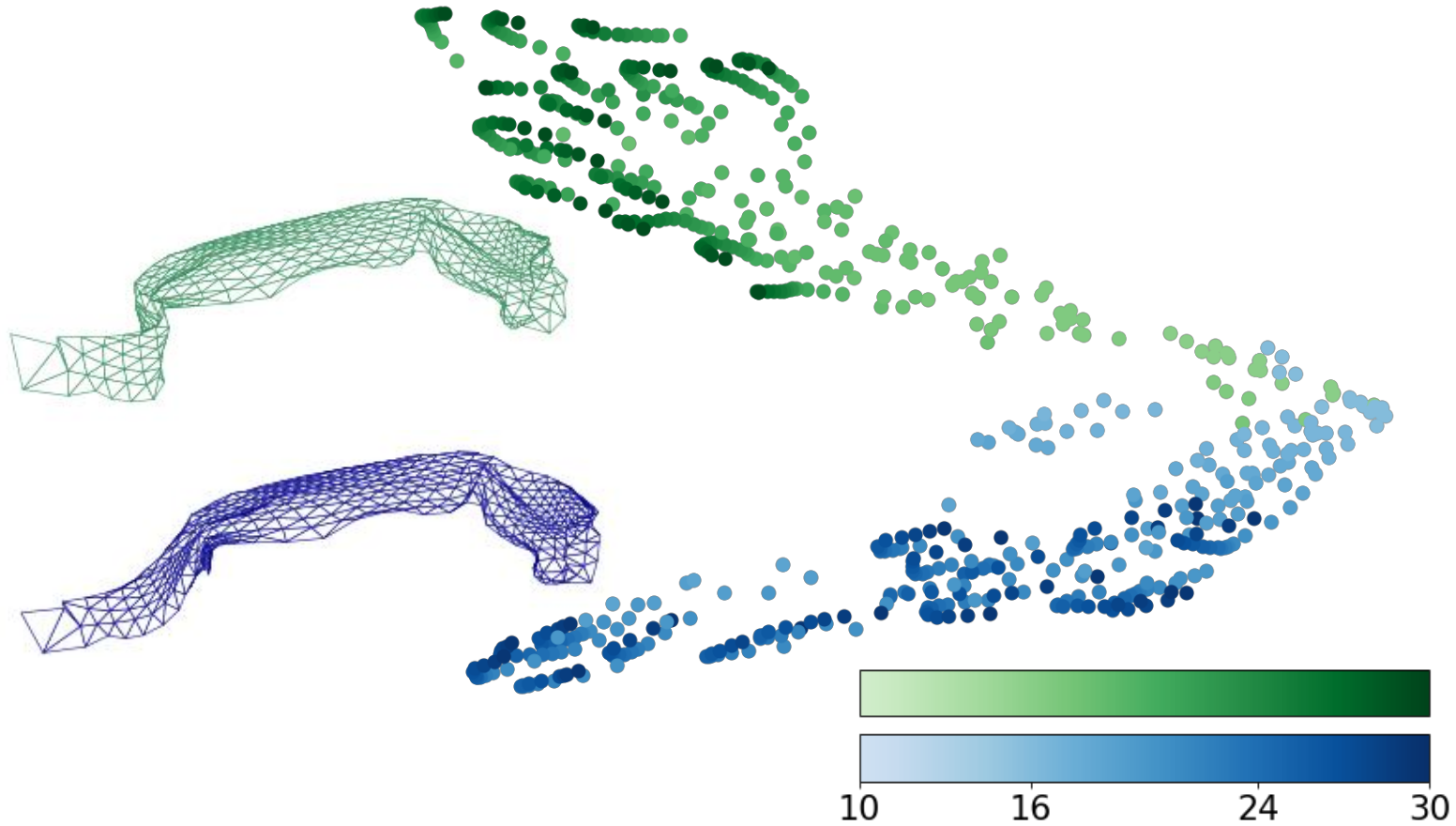}
\end{center}
\caption{Embedding of the TRUCK's left front beam for $t=10,\dots,30$. 32 simulations deform in two branches. 
Color encodes timestep and branch.}
\label{emb_truck}
\end{figure}

Figure \ref{emb_truck} shows the embedding in the low-dimensional space for the TRUCK's left front beam. The beam deforms in two different branches, which manifests in the embedding. The results are similar to \cite{Bohn2013,Hahner2020}.

\begin{figure}[!h]
    \centering
    {\includegraphics[width=0.8\linewidth, trim=17cm 12.8cm 9cm 12cm, clip]{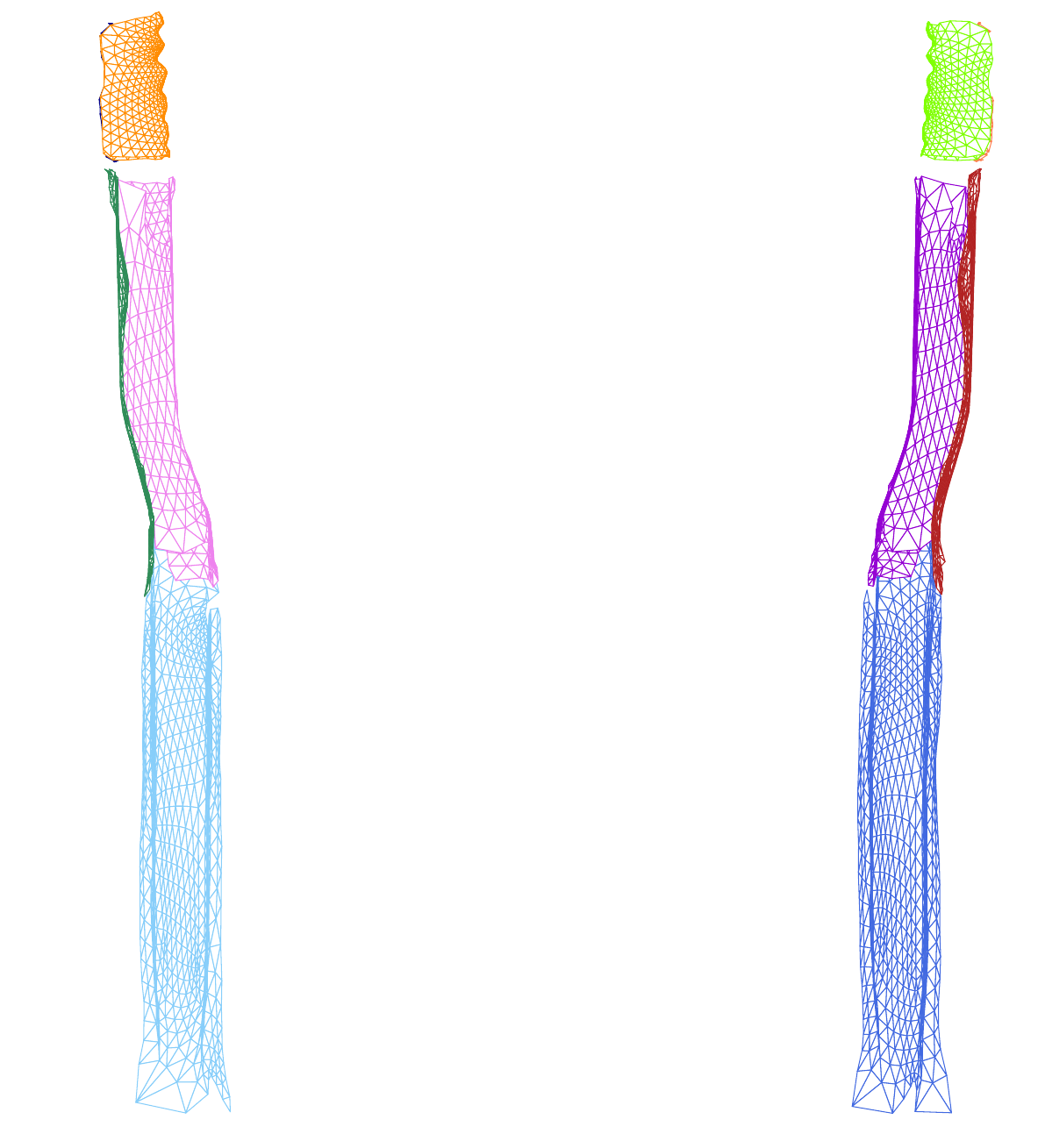}    }
    \caption{Selected car components in the YARIS dataset.}
    \label{yarisparts}
\end{figure}

\begin{figure}[!h]
    \centering
    {\includegraphics[width=0.7\linewidth, trim=23.5cm 12.8cm 9cm 12cm, clip]{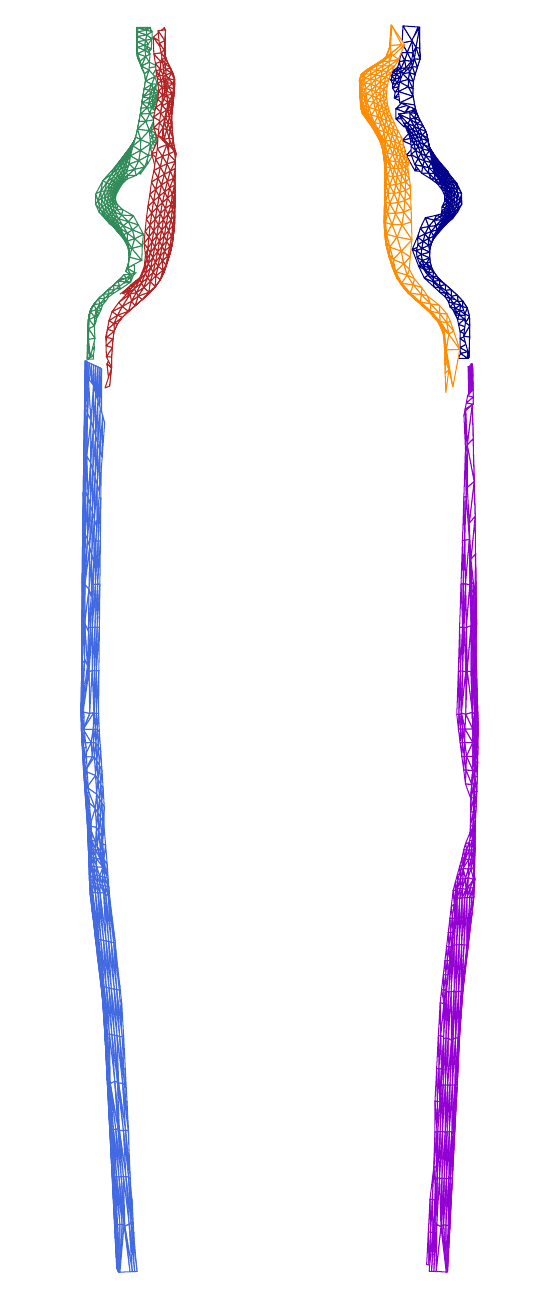}    }
    \caption{Selected car components in the TRUCK dataset.}
    \label{truckparts}
\end{figure}

For the YARIS and TRUCK dataset we visualize in Figure \ref{yarisparts} and Figure \ref{truckparts} respectively the selected car components, whose deformation over time we analyze with the mesh autoencoder for semi-regular meshes. All the components have different mesh representations, which we handle with only one autoencoder.

\end{document}